\documentclass[11pt,letterpaper]{article}
\usepackage{amsmath}
\usepackage{amsfonts}
\usepackage{amssymb}
\usepackage{mathrsfs}
\usepackage{graphicx}
\usepackage[margin=1in]{geometry}
\usepackage[hidelinks]{hyperref}
\usepackage{multirow}
\usepackage{pifont}
\usepackage{float}
\usepackage{natbib}
\usepackage{tipa}
\usepackage{tikz}

\title{Deep Sound Change: Deep and Iterative Learning, Convolutional Neural Networks, and Language Change}
\author{Ga\v{s}per Begu\v{s}\\Department of Linguistics, University of California, Berkeley\\\texttt{begus@berkeley.edu}}

\date{}

\begin{document}


\maketitle


\begin{abstract}

This paper proposes a framework for modeling sound change that combines deep learning and iterative learning. Acquisition and transmission of speech is modeled by training generations of Generative Adversarial Networks (GANs) on unannotated raw speech data. The paper argues that several properties of sound change emerge from the proposed architecture. GANs \citep{goodfellow14,donahue19} are uniquely appropriate for modeling language change because the networks are trained on raw unsupervised acoustic data, contain no language-specific features and, as argued in \citet{begus19}, encode phonetic and phonological representations in their latent space and generate linguistically informative innovative data. The first generation of networks is trained on the relevant sequences in human speech from TIMIT. The subsequent generations are not trained on TIMIT, but on generated outputs from the previous generation and thus start learning from each other in an iterative learning task. The initial allophonic distribution is progressively being lost with each generation, likely due to pressures from the global distribution of aspiration in the training data. The networks show signs of a gradual shift in phonetic targets characteristic of a gradual phonetic sound change. At endpoints, the outputs superficially resemble a phonological change --- rule loss.

\textbf{keywords}: deep learning, iterative learning, historical linguistics, generative adversarial networks%
\end{abstract}

\section{\label{intro} Introduction}

Sound change is one of the most widely studied phenomena of human language and behavior in general \citep{h08,g14}.  It has long been established that languages change constantly on all levels: phonetic, phonological, morphological, syntactic, and semantic. For example, Modern French developed from (Vulgar) Latin. The two languages differ in phonemic inventories and their phonetic realizations, morphology, word order, and in individual words meanings. Language change on the phonetic/phonological level (i.e.~sound change) is arguably the most systematic and well-studied. Despite the relatively long history of scientific inquiry into the topic \citep{paul80}, many open questions remain regarding the causes of sound change and language change in general. Even its definition is problematic, as \citet{g14} notes: ``[d]espite its ubiquity, there is no generally accepted definition of sound change.''

\subsection{Prior work}

Due to the lack of direct empirical evidence on how exactly sound changes arises, computational simulations have been used as substitutes to better understand mechanisms behind sound change. Several computational models have been proposed that simulate agents (speakers) learning some linguistic input and interacting with other agents. Typically, over the course of generations of such agent-based learning, a change in some phonetic feature occurs which replicates sound change in progress \citep{baker08}. For example, peripheral vowels [a], [i], and [u] reduce to a central vowel [\textipa{@}] when agents with a bias for reduction interact in an iterative learning task \citep{deboer03}. By comparing computational models with properties of sound change that are empirically observed, we can evaluate models and the degree to which they approximate reality. If the model matches observed behavior in sound change, it can be argued that the assumed architecture of the model is responsible for the match, which means that the relevant mechanisms likely influence the actual sound change.

One of the prominent frameworks in computational simulations of sound change is the exemplar theory model \citep{johnson97,p01,wedel06,kirby15,duran17} which assumes speakers store some perceived phonetic information (such as targets for vowel formants F1--F4) as ``remembered tokens" or exemplars. Phonetic categories consist of a set of  such exemplars, grouped by ``similarity across any salient dimension'' \citep{wedel06}. Categories are updated with each new token and each new production samples from this distribution, creating a range of possible values within the exemplar, which thus introduces phonetic variation in speech production.  Several agent-based simulations of sound change have been proposed in which agents interact and learn from each other some abstract and extracted phonetic representation \citep{deboer03,kirby15}.

The existing computational simulations of sound change require strong assumptions about learnability and a high level of abstraction of phonetic features. First, the models need to assume a simplified representation of learning: agents are already assumed to be able to learn and store phonetic features and values. Second, the models are not trained on raw acoustic inputs, but on items that only tangentially represent linguistic reality\footnote{Earlier simulations operated with fully abstract symbols, such as $p$ and $q$ that represent two variants in \citet{nettle99}.} or are extracted and reduced forms of the phonetic signal. For example, learning of vowels is usually modeled with absolute values of formats F1--F4. When an agent observes or learns a vowel, it stores the representation reduced to the four dimensions. Occasionally, noise and bias are added, but the learning in the model is still  highly abstracted and does not closely resemble acquisition in language learning infants or even speech perception in adults. Recently, more elaborate models have been proposed, but the dimensions that the agents learn are still highly abstracted; for example, absolute formant values \citep{wedel06,kirby15,stevens19}. Hearing language-acquiring children need to learn articulatory targets for vowels and other sounds from raw acoustic inputs and not from already extracted simplified dimensions.

Neural network architectures have long influenced computational models of language. The majority of neural network approaches to modeling language data operate on a morphological or syntactic levels with fewer applications in phonetics and phonology. Nevertheless, neural networks have been widely use to model phonetic learning both in the earlier works \citep{mccleland86,gaskell95,plaut99,kello03}  as well as more recently with autoencoder  and other deep learning architectures \citep{rasanen16,eloff19,shain19,begus19,begusCiw}.

Populations of neural networks have also been used to model language change  in the earlier development of neural network research (for an overview, see \citealt{baker08}). For example, \citet{hare95} model morphological change with populations of neural networks. Training is stopped early in the model to simulate imperfect learning. After generations of networks, the first of which is trained on Old English data (whereby subsequent generations start learning from each other), the morphological system becomes more regular, reflective of Modern English \citep{hare95,baker08}. 
Likewise, \citet{livingstone02}  trains fully connected two-layer neural networks  to model emergence of dialects. However, these models require highly abstracted representations, are not trained on raw acoustic data, and are not fully generative in the sense that the networks would output innovative data. To the author's knowledge, none of the proposals within the deep neural network framework model sound change in an iterative learning setting trained on raw speech.

\subsection{Goals}
  
This paper models language change with deep convolutional neural networks within the Generative Adversarial Networks framework \citep{goodfellow14} that are trained on raw acoustic inputs. \citet{begus19} argues that acquisition of phonetics and phonology can be modeled as a dependency between latent space and generated data in the GAN framework and proposes a technique to identify the network's internal learning representations that are phonetically and phonologically meaningful. To model sound change, we introduce GANs to the iterative learning paradigm \citep{kirby01}: agents in the model are individual Generative Adversarial Networks (as implemented for audio data in \citealt{donahue19} based on Deep Convolutional Generative Adversarial Networks by \citealt{radford15}), which learn to output acoustic data that resembles speech of the raw acoustic data they are trained on. 

The following iterated learning architecture is proposed: the initial input for the first generation of neural networks (a single GAN) are sequences of sounds from the TIMIT database \citep{timit}. The network is trained on this input and after some number of epochs, generates data that closely resemble the input speech. The next generation of networks is trained on the generated data from the previous generation. This is iterated as long as the networks output data that can be modeled with the standard acoustic methodology for speech data. This iteration between generations of networks superficially simulates the transmission of speech between actual language learners and provides the opportunity to study sound change in progress through a process that mirrors real language learning and transmission. 

We tests the development of an allophonic distribution based on English primary data. In English, stops [\textipa{p\super h}, \textipa{t\super h}, \textipa{k\super h}] are generally aspirated --- produced with a long Voice Onset Time (VOT) or puff of air ---  before stressed vowels in absolute word-initial position (when no other consonant precedes the stop) and unaspirated [p, t, k] if an [s] precedes \citep{lisker84,iverson95,vaux02,vaux05,davis06,ahmed20}, e.g.~[\textipa{"pIt}] `pit'  and  [\textipa{"spIt}] `spit'). English L1 learners thus need to acquire long VOT (aspiration) and shorten it if an [s] precedes.   Regardless of how we formalize this process (in symbolic rewrite rules as in \citealt{ch68} or connectionist constraints in \citealt{ps93}), the learner needs to acquire a conditional distribution: short VOT if [s] precedes, otherwise long VOT. If the acquisition is imperfect, the learner fails to shorten the VOT in the [s]-condition and extends the unmarked \emph{elsewhere} variant  --- long VOT --- into the more restricted [s]-condition. This stage is confirmed by language acquisition research: children acquiring English as L1 produce longer VOT and occasionally very long VOT durations in the s-condition compared to the adult population \citep{bond80,bond81,mcleod96}, likely because they fail to acquire the distribution perfectly. Similarly, patients with speech impairments produce long VOT in the [s]-condition because they fail to apply the phonological rule \citep{buchwald12}.

By training generations of GANs on a minimal allophonic distribution, we test which predictions of language change can emerge in deep convolutional neural networks. First, we test whether GANs that learn from each other can output linguistically analyzable data.  We illustrate that acoustic phonetic analysis of both the duration (Section \ref{VOTduration}) and spectral properties (Section \ref{spectral}) can be performed on the generated outputs and that such an analysis yields informative results. Second, we test whether generations of GANs are able to represent properties of the data with the same frequencies as in the original training distributions (Section \ref{proportion}). For example, we analyze how well the less frequent \#sTV (where \# marks the beginning of the work, T a voiceless stop, and V a vowel) sequences are represented relative to TIMIT.  As already mentioned, both language-learning children and convolutional neural networks show evidence of imperfect learning of the aspiration rule in English --- they occasionally fail to apply the rule. We test the effects of such imperfect learning on outputs of deep convolutional networks in iterative learning tasks. In other words, can failure to fully acquire a rule accumulate in the loss of a rule?  Imperfect acquisition predicts that the more general \#TV condition will remain unchanged, while the restricted \#sTV condition  will undergo loss of a distribution, which resembles phonological pressures in natural language.  We argue that appearance of such a loss indeed emerges in GANs trained iteratively and  that gradual shifts of phonetic targets underlie the loss of a distribution (Section \ref{discussion}). 

The advantage of the proposed model is that the networks are fully generative and unsupervised: they need to learn speech data from noise in the latent space based on raw acoustic inputs in the training data, replicating the conditions of infant language acquisition. The model does not require any level of abstraction; the only data manipulation that the models require is slicing acoustic inputs into sequences of sounds. No assumptions about learning are necessary and no language-specific features are assumed in the model. 

Additionally, neural networks in the GAN architecture do not access the training data directly: they need to learn to output data from noise. \citet{begus19} argues that GANs produce innovative outputs that are structured, highly informative, and consistent with linguistic behavior in language acquisition, speech impairments, and speech errors. \citet{begus19} also argues that the networks learn to use the random space as an approximation to phonetic and phonological features. Several parallels between phonetic and phonological learning in language-acquiring children and neural networks are thus observed. This paper tests whether similarities emerge also when Generative Adversarial Networks learn from each others in an iterative learning architecture.

The proposed experiment also has implications for the discussion on distinguishing different influences on sound change. While the proposed model does not include any articulatory or perceptual phonetic forces, diverging from modeling reality, it is in fact a desirable property for computational tests of different hypothesis about sound change. A model without articulatory information can provide evidence for changes that do not require articulatory bias, thus controlling for one factor in the set of possible influences on sound change. The Generative Adversarial Neural network architecture also lacks any domain-specific linguistic elements. This means that the proposed model can test which properties of sound change can be derived with domain-general cognitive mechanisms and transmission in absence of language-specific or articulatory phonetic influences.
Combining iterative learning with deep convolutional networks trained on raw audio data is also part of a broader goal to test which properties of language and language acquisition can be derived with deep convolutional networks in the GAN setting.

\section{Materials}

\subsection{Background: Generative Adversarial Phonology}

The GAN architecture first proposed by \citet{goodfellow14}  involves two networks, a Generator network (G) and a Discriminator network (D). The Generator is trained to map latent random variables ($z$) to data, while the Discriminator is trained on distinguishing real data ($x$) from generated data ($\hat{x}$). The Generator is trained to maximize the Discriminator's error rate, and the Discriminator is trained to minimize its own error rate. This minimax training results in a Generator network that outputs data that are increasingly difficult to distinguish from real data.

\citet{radford15} introduce deep convolutional networks to the GAN architecture (DCGAN). The DCGAN model has recently been  transformed from modeling two-dimensional visual data to one-dimensional time-series audio data in a proposal called WaveGAN (used in this experiment) by \citet{donahue19}. The architecture includes the Generator network as a five-layer deep convolutional network that takes as its input 100 latent $z$-variables distributed uniformly on the interval $z_{1-100}\sim\mathcal{U}(-1,1)$ and outputs 16,384 data points that constitute over 1 s of audio file (sampled at 16 kHz). The Discriminator similarly takes as an input 16,384 data points of either generated ($\text{G}(x)$) or real ($x$) data and estimates Wasserstein distance according to the proposal in \citet{arjovsky17}. For a detailed description of the hyperparameters, see \citet{radford15,donahue19}.

\citet{begus19,begusIdentity,begusCiw,begusLocal} argues that deep convolutional networks in the GAN setting learn to encode  phonetic and phonological representations in the latent space. The network is trained on the same allophonic distribution as in this paper (using the same data), but only for one generation. \citet{begus19} showed that the network can successfully learn a simple allophonic distribution, such as the English aspiration distribution ([\textipa{"p\super hIt}] $\sim$ [\textipa{"spIt}]). The advantage of modeling speech acquisition with GANs is that the networks not only replicate data, but generate innovative outputs. The innovative outputs are highly informative and linguistically interpretable. For example, the Generator learns to recombine phonetic units into unobserved novel combinations that can be analyzed as linguistically valid. A model trained on \#TV and \#sTV sequences (such as [\textipa{"t\super h\ae}] and [\textipa{"st\ae}]) can output \#TTV or \#sV sequences that were not part of the training data \citep{begus19}. Similarly, in a model architecture for lexical learning fiwGAN (proposed in \citealt{begusCiw}), the Generator outputs \emph{start} [\textipa{"stA\*rt}], although  [\textipa{"stA\*rt}] was not part of training data: the network is only trained on \emph{dark}, \emph{suit} and six other unrelated words.

The Generator also learns to use a subset of latent variables to represent approximates of phonological concepts such as presence of [s]. \citet{begus19} proposes a technique to identify latent variables that correspond to some phonetic or phonologically meaningful representation and argues that the relationship between the variables and presence of the representation in the output is near linear. By manipulating individual variables to values well outside the training interval  \citep{begus19}, we can actively force presence or absence of a segment such as [s] in the output. Interpolating a single variable results in gradual increase or decrease of frication noise of [s] \citep{begus19}. While manipulating internal representations of the trained networks is highly informative for probing tasks, no variables are manipulated in the iterative learning experiments presented here --- the objective of the paper is  to test what types of changes emerge in an iterative learning setting without any active manipulations to the networks.

\subsection{Model}

\begin{figure}
\centering
\includegraphics[width=1\textwidth]{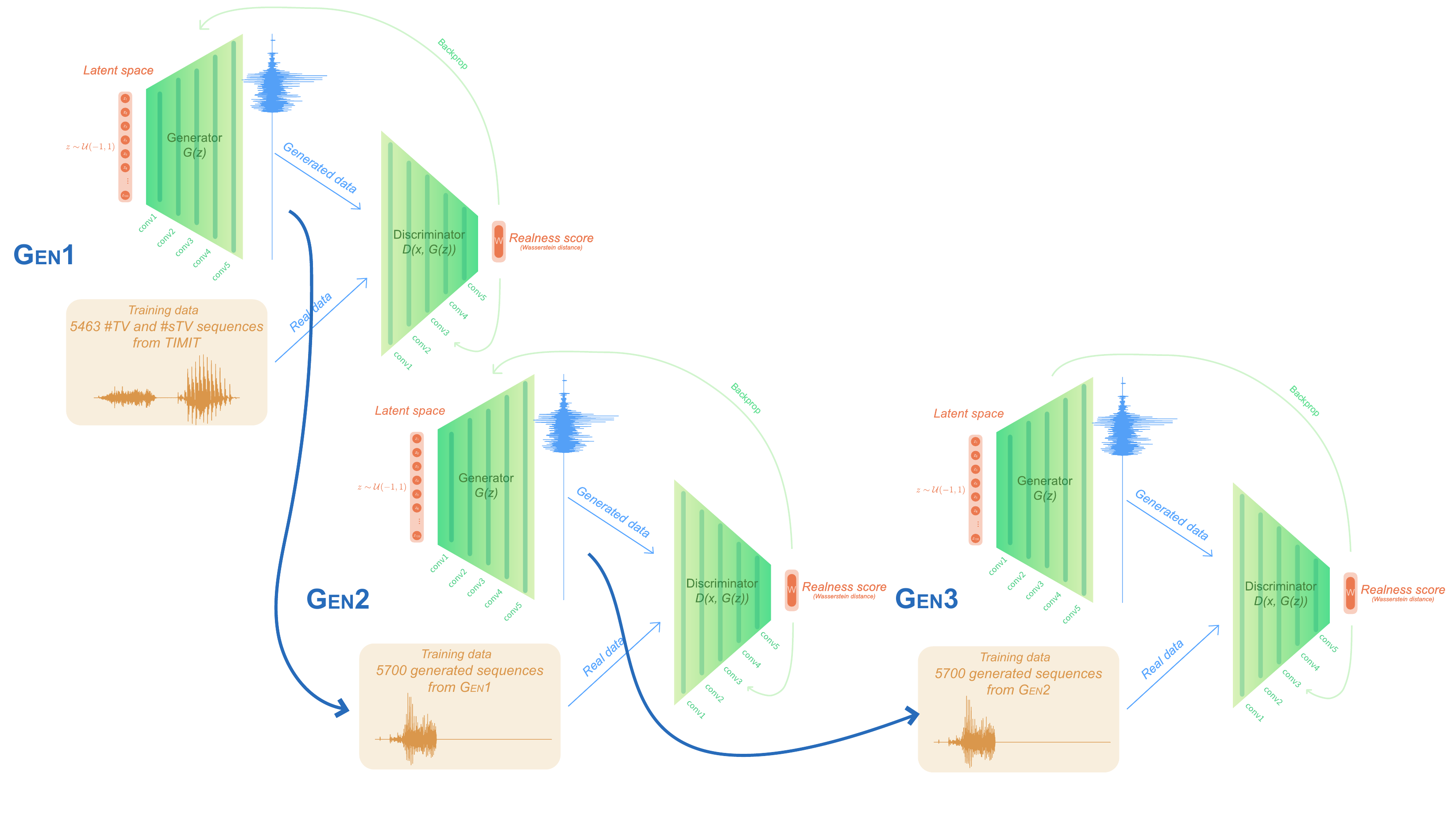}
\caption{\label{tikzflowSC} A diagram of generations of Generative Adversarial Networks as trained in this paper (each network is trained on the proposal in \citealt{donahue19}). Gen 4 is omitted from the figure. The tearm ``realness'' is taken from \citet{brownlee19}.}
\end{figure}

The computational experiment is performed on WaveGAN \citep{donahue19}, which is an implementation of the DCGAN architecture \citep{radford15} for audio data with few modifications. Figure \ref{tikzflowSC} illustrates the training procedure. The first network (Gen1) is trained on data from the TIMIT database \citep{timit}. The training data consists of 4,930  sequences with the structure \#TV and 533  sequences with the structure \#sTV (5,463 total). The training is stopped after 12,255 steps. At this point, 5,700 outputs are generated from \textsc{Gen1}. No manipulations are performed on the latent space. These 5,700 data points constitute the training data for the second network: \textsc{Gen2}. \textsc{Gen2} is thus trained exclusively on the generated outputs of the first network, \textsc{Gen1}. After 12,239 steps, the training is stopped, at which point the Generator network outputs 5,700 data points that are fed to the next network. Altogether four generations of networks were thus trained (\textsc{Gen1, \textsc{Gen2}, \textsc{Gen3}, \textsc{Gen4}}). All networks are trained on approximately equivalent number of training steps and data points. Because the training is stopped at a relatively small number of training steps,  noise arises with each generation of networks to induce imperfect learning. For this reason, we only trained four generations. Table \ref{gentable} lists training steps and details about data points of the four networks. The networks were trained at an approximate pace of 25 steps per 300 s on an  NVIDIA Tesla M10 GPU unit.

\begin{table}
\centering
\begin{tabular}{lrrrr}
\hline\hline
&Steps&Epochs &Training data&Source of data\\\hline
\textbf{Gen1}&12,255&717&5,463&TIMIT (=\textsc{Gen0})\\
\textbf{Gen2}&12,239&687&5,700&Gen1\\
\textbf{Gen3}&12,249&687&5,700&Gen2\\
\textbf{Gen4}&12,246&687&5,700&Gen3\\

\hline\hline
\end{tabular}

\caption{\label{gentable}Generations of trained networks, their training steps, and number of data points in training.}
\end{table}

\subsection{Data}

The data used for training (\textsc{Gen0})  in the first generation (\textsc{Gen1}) are sliced from the TIMIT database \citep{timit}. The TIMIT database includes 10 sentences per each of the 630 speakers (total 6,300 sentence) from 8 dialectal groups, which provides a relatively diversified corpus.   For the purpose of training the networks on an allophonic distribution, sequences with the structure \#TV and \#sTV were sliced from the database.\footnote{The data and the first network (\textsc{Gen1}) are the same as in \citet{begus19}.} In other words, the networks are trained on TIMIT sequences of a voiceless stop /p, t, k/ (represented by T) at the beginning of a word (represented by \#), followed by a vowel (represented by V)\footnote{TIMIT uses the ARPABET convention. The following vowels are included in the analysis (in ARPABET and IPA transcriptions): aa = \textipa{A}, ae = \textipa{\ae}, ah = \textipa{\textturnv}, ao = \textipa{O}, aw = \textipa{aU}, ax = \textipa{@}, ax-h = \textipa{\r*{@}}, axr = \textipa{\textrhookschwa}, ay = \textipa{aI}, eh = \textipa{E}, er = \textipa{\textrhookrevepsilon}, ey = \textipa{eI}, ih = \textipa{I}, ix = \textipa{1}, iy = \textipa{i}, ow = \textipa{oU}, oy = \textipa{oI}, uh = \textipa{U}, uw = \textipa{u}, ux = \textipa{0}.} and optionally preceded by a sibilant fricative [s] (represented by s). The training data are limited to vowels and do not include any other segments after the vowel in order to control for unwanted effects of other potential conditional distributions \citep{begus19}. Vowels are distributed  approximately equally across the two conditions:  \#TV and \#sTV.

\begin{figure}
\centering
\includegraphics[width=0.45\textwidth]{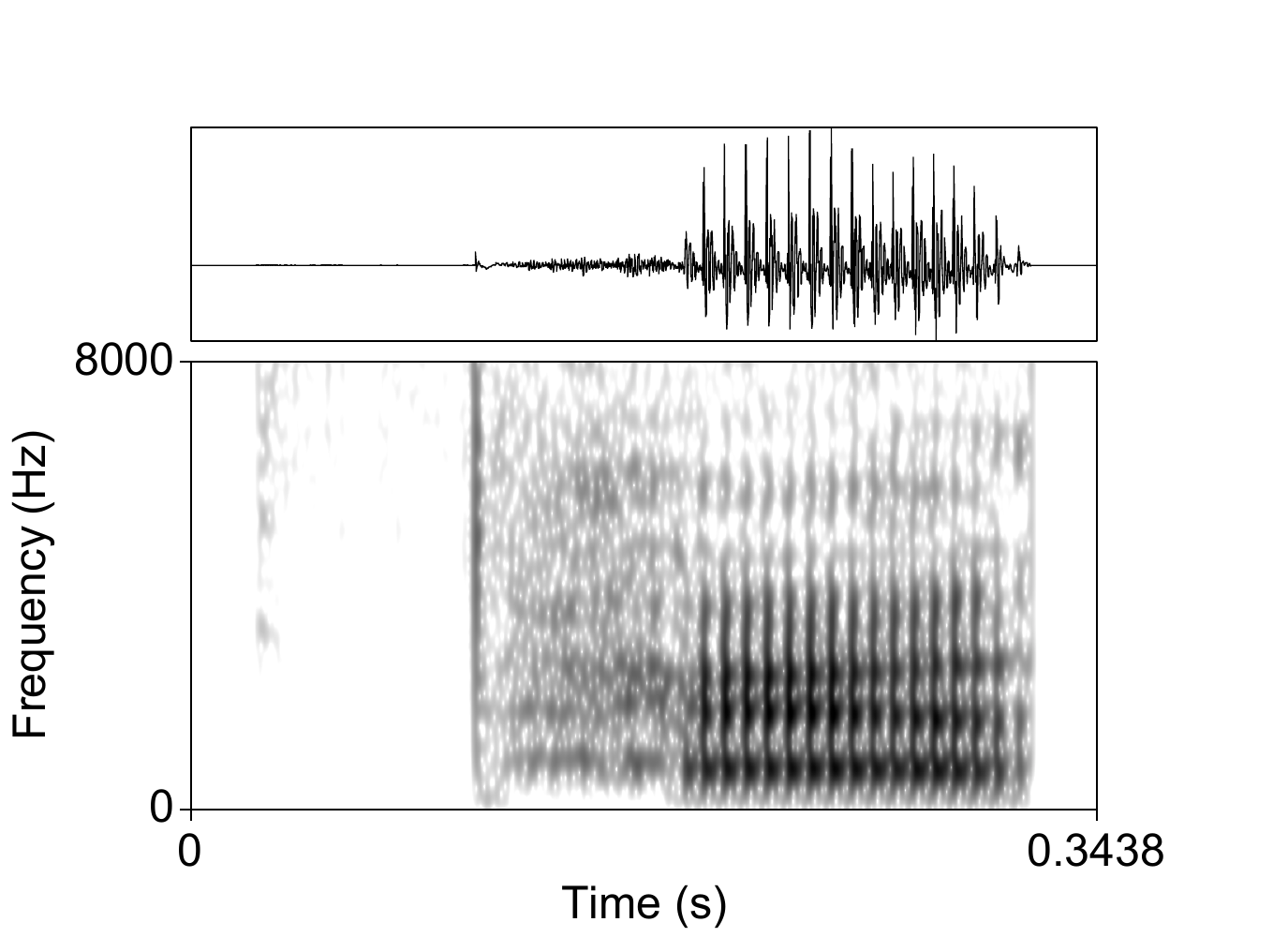}\includegraphics[width=0.45\textwidth]{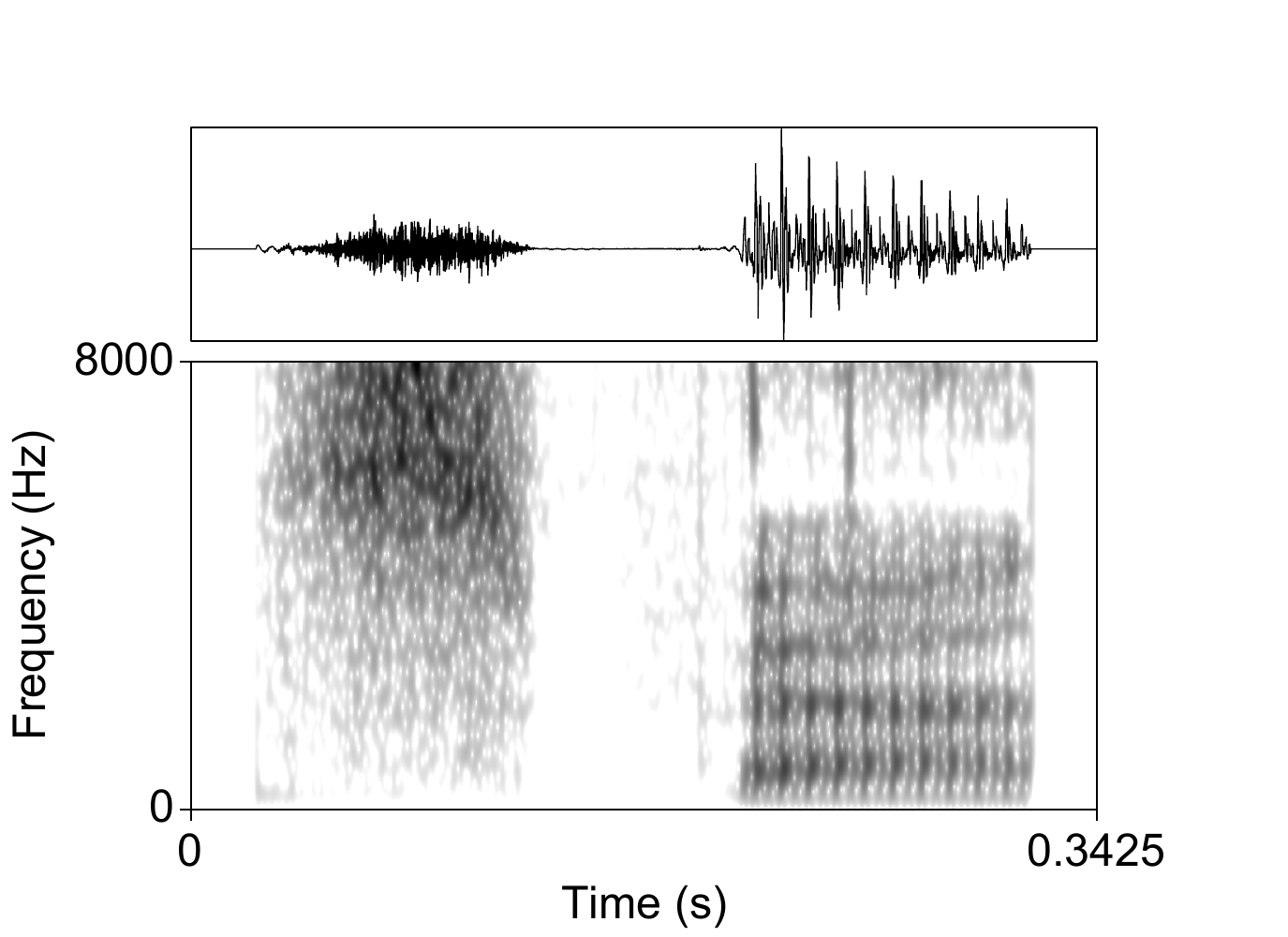}
\caption{\label{artificialSCspae} Waveforms and spectrograms (0--8,000 Hz) of a typical \#TV and a \#sTV sequence ([\textipa{p\super h\ae}] and [\textipa{sp\ae}]) from the TIMIT databased used for training. The networks are trained on waveforms exclusively: spectrograms are given here only for illustration. Waveforms in training are additionally padded with silence until they reach 16,384 datapoints.}
\end{figure}

The sequences are raw acoustic waveform files, sampled at 16 kHz and fed to the model (which pads them with silence into exactly 16,384 data points that constitute the waveform).  The training data are raw waveform files and include no segmentation or any other information. The only supervision in the model is the fact that data is sliced into sequences of the structure \#TV and \#sTV. Figure \ref{artificialSCspae} illustrates two typical training sequences sliced from TIMIT, one for each of the two conditions.

\section{Results}

\subsection{Proportion of \#sTV sequences}
\label{proportion}

To analyze learning of the allophonic distribution across generations in deep neural networks, we first generate 2,000 outputs from the Generator network for each of the four generations of networks (8,000 outputs total) and test how the proportion of \#sTV sequences changes with each generation. 

The TIMIT training data for \textsc{Gen1} contains altogether 5,463 sequences, 533 of which are of the \#sTV structure. In other words, 9.76\% of the initial training data are \#sTV sequences. This ratio appears slightly lower in the next generations, but not substantially so.\footnote{All annotations in the paper are performed by the author. The difference between outputs without [s] and those containing an [s] are acoustically very salient. To ensure consistency of annotations for  less salient properties (such as annotating VOT), the author manually analyzed the data several times and made data and annotations available.} Of the 2,000 outputs examined per each generation, the number of \#sTV sequences range from  119 to 157 (or 5.95\% to 7.85\% see Table \ref{gentableprop}).

\begin{table}
\centering
\begin{tabular}{rrrrr|r}
\hline\hline
&All sequences&\#sTV &\#TV&\%of \#sTV&\#TV annotated\\\hline
\textbf{(TIMIT) \textsc{Gen0}}&5,463&533&4,930&9.76\%&4,930\\
\textbf{Gen1}&2,000&151 &1,849&7.55\%&157\\
\textbf{Gen2}&2,000&153 &1,847&7.65\%&157\\
\textbf{Gen3}&2,000&157&1,843&7.85\%&157\\
\textbf{Gen4}&2,000&119&1,881&5.95\%&157\\

\hline\hline
\end{tabular}

\caption{\label{gentableprop}Counts of analyzed data and proportion of \#sTV in the output data across the generations. \textsc{Gen0} represents the TIMIT database.}
\end{table}

To test whether the differences in the proportion of \#sTV sequences across the four generations and the TIMIT initial training data are significantly different, we fit the counts of \#sTV and \#TV sequences from Table \ref{gentableprop} to a negative binomial regression model. The dependent variable is the count of the sequences; the independent variables are \textsc{Generations} (treatment-coded with \textsc{Gen1} as the reference level) and \textsc{Structure} (\#sTV vs.~\#TV, treatment-coded with \#TV as reference) with the interaction.\footnote{The regression is performed with the \emph{gam()} function in \emph{mgcv} \citep{mgcv} in R statistical software \citep{r}. } 

\begin{figure}
\centering
\includegraphics[width=0.6\textwidth]{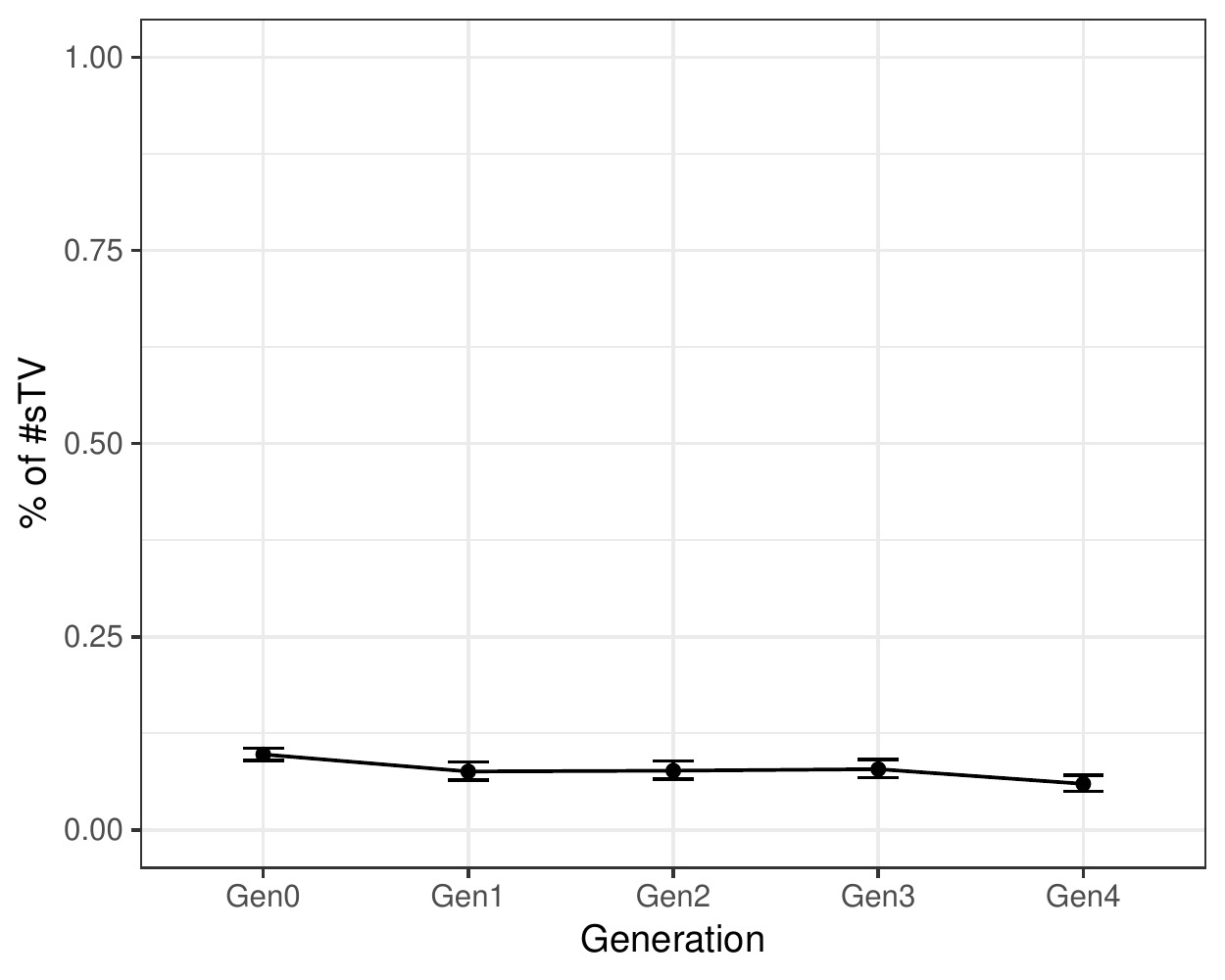}
\caption{\label{evolutionGANproportionS} Proportion of \#sTV sequences in the output across generations of trained network when data is fitted to a linear logistic regression model (not a negative binomial model). \textsc{Gen0} represents the proportion in the TIMIT database on which the first generation (\textsc{Gen1}) is trained. }
\end{figure}

Overall, the proportion of \#sTV sequences in the output is significantly smaller in the first generation  (\textsc{Gen1}) compared to the training data from TIMIT (\textsc{Gen0})  ($\beta=0.28,z= 2.84, p=0.005$) and slightly smaller in \textsc{Gen4} than in \textsc{Gen1} ($\beta=-0.26,z=-1.98,p=0.048$). The differences between \textsc{Gen1} and \textsc{Gen2} and \textsc{Gen3} are not significant (full model in Table \ref{gentablepropLogReg}). Figure \ref{evolutionGANproportionS} illustrates the distribution (estimates are from a logistic regression model). It appears that the proportion of the \#sTV sequences compared to the \#TV sequences is relatively stable across the generations of networks.

\subsection{VOT duration}
\label{VOTduration}

To evaluate how learning of the allophonic distribution changes with each generation, we compare VOT durations in \#TV and \#sTV sequences across the four generations and TIMIT. The author annotated all \#sTV sequences in the 8000 generated outputs (4 generations $\times$ 2000) and a number of \#TV sequences that equaled the highest number of \#sTV sequences in the outputs per generation, i.e.~157.\footnote{The annotated \#TV sequences for VOT duration measurements  were chosen from the beginning of the 2,000 generated data samples in order to represent the durations of \#TV sequences consistently across the generations and to reduce selection bias.} Measurements of VOT durations in the TIMIT database used for training the first generation, are taken from the 5,463 TIMIT-internal hand-annotations.

Identifying \#sTV sequences and especially  boundaries between aspiration duration and closure/onset of the vowel in the generated outputs is a challenging task. Since automatic speech tagging systems are still underperforming trained human phoneticians even for acoustically cleaner and less ambiguous data, human annotation is necessary for data annotations for the current levels of audio quality in the generated data. The author manually inspected all 8,000 data points. In order to keep annotation criteria constant across the generations, the entire dataset (8,000 outputs) were inspected four times. 

\begin{figure}
\centering
\includegraphics[width=0.4\textwidth]{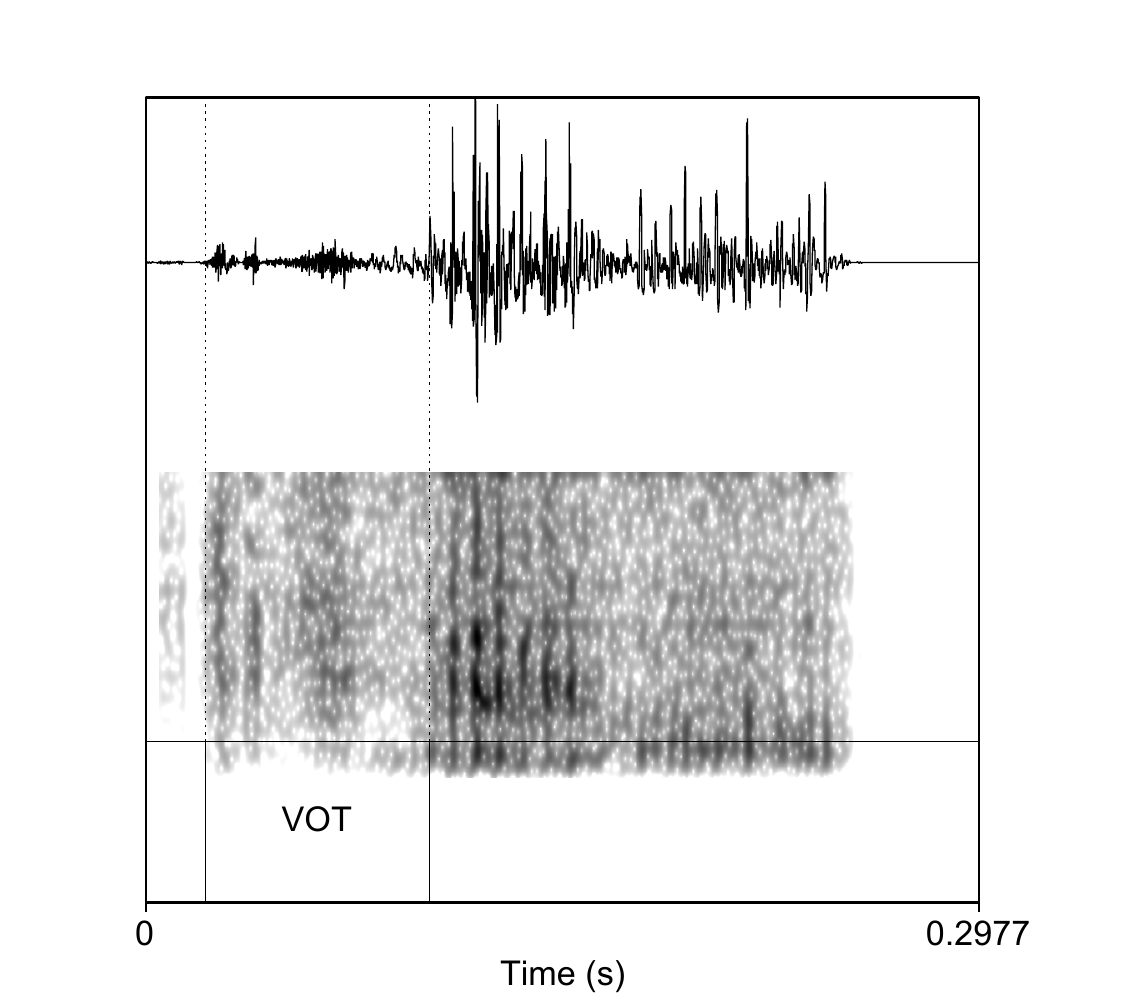}
\includegraphics[width=0.4\textwidth]{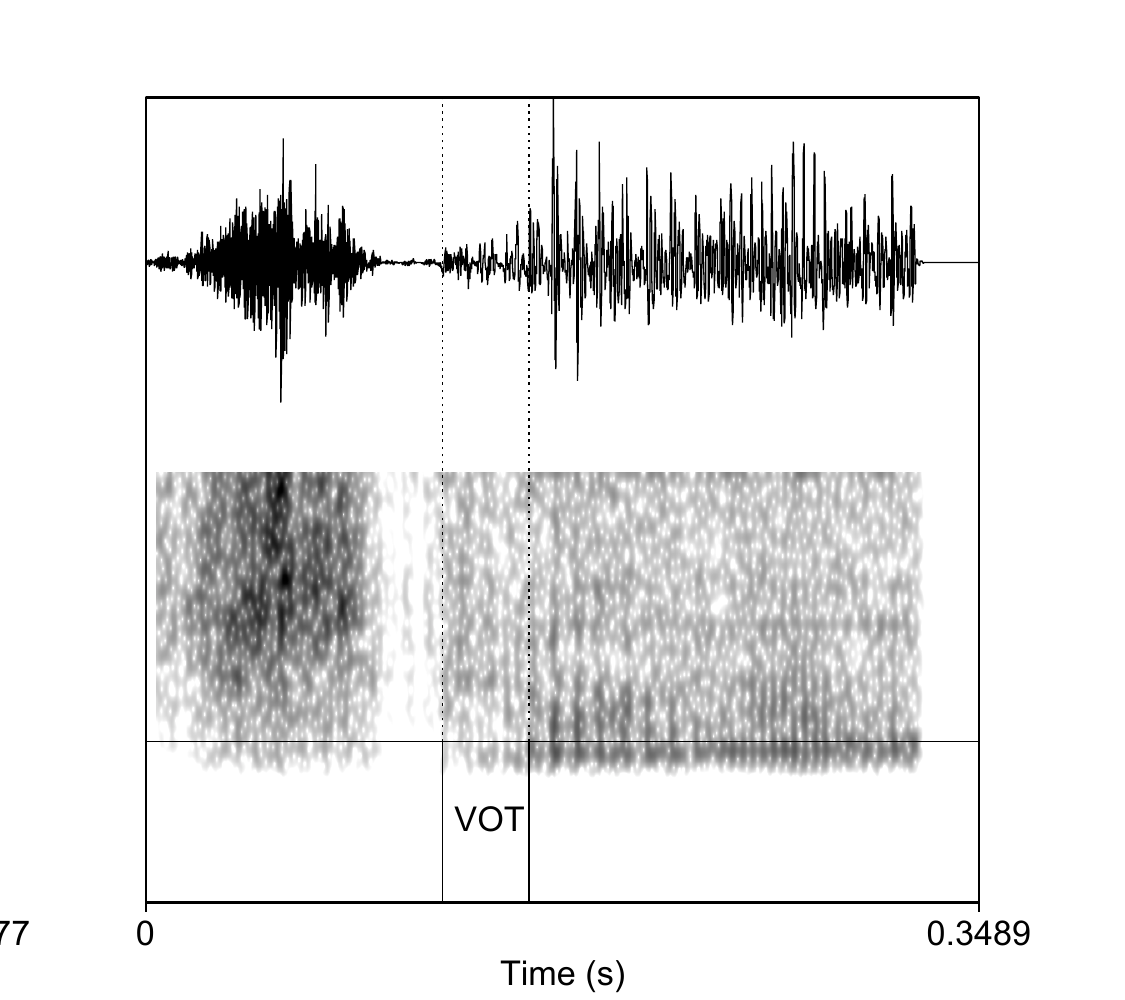}
\caption{\label{ASCannotation} Waveforms and spectrograms (0-8000 Hz) of two generated outputs from \textsc{Gen4} with corresponding annotations.}
\end{figure}

The generated outputs are not recordings of actual  human speech, which is why finding boundaries between the release of the following stop and onset of the vowel is particularly challenging. Two example waveforms and spectrograms with annoatations are given in Figure \ref{ASCannotation}. The main criteria for phonetic annotation were consistency across generations and finding acoustic equivalents of closure, stop release, aspiration noise, and vowel onset. Borderline cases analyzed as either \#sTV (and therefore included in the analysis) or \#sV/\#TV (and therefore excluded) were decided based on consistency of criteria across generations. Guidelines for data annotations as described in \citet{ladefoged03} were used.\footnote{Generated data, annotations, and trained checkpoints from the four models are available here: \url{https://doi.org/10.17605/OSF.IO/5KBEF}.} 

 The main question of the present experiment is whether the difference in aspiration between the two conditions changes significantly across generations of deep convolutional neural networks. While the most prominent phonetic correlate of aspiration is VOT duration and spectral properties of burst and aspiration noise \citep{klatt75,hussain20}, the duration of aspiration (VOT)  is influenced by many other phonetic factors (such as place of articulation or vowel quality). However, in the Generated data, even VOT durations, which are phonetically prominent, are occasionally challenging to annotate. Vowel quality and place of articulation are very difficult to reliably analyze. For this reason, we analyze VOT durations and the spectral properties of the aspiration noise (Section \ref{spectral}). To control for other influences, we analyze  duration and spectral properties in the \#sTV condition \emph{relative to} the \#TV condition.  Given that we do not expect distributional differences in other phonetic parameters across the generations, the results should not be substantially impacted by other phonetic features. Additionally, the spectral analysis of frication noise  in Section \ref{spectral} suggests that no substantial changes in spectral properties occur across the four generations.

\begin{figure}
\centering
\includegraphics[width=0.7\textwidth]{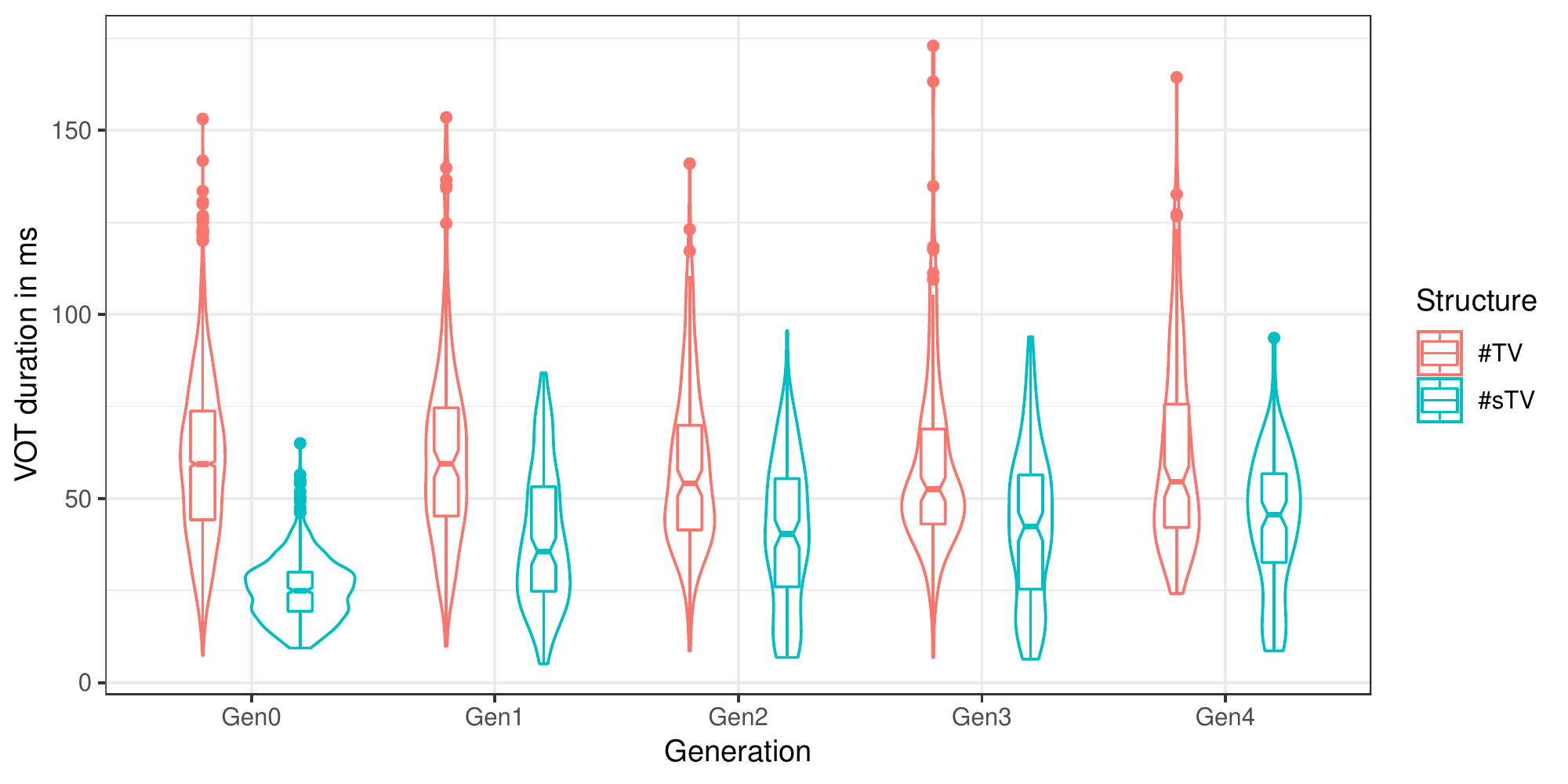}
\caption{\label{evolutionGANviolin} Violin and boxplot of VOT duration in ms across the four generations of trained networks and TIMIT.}
\end{figure}

The raw data suggest that the difference in VOT durations between the two conditions, \#TV and \#sTV, becomes increasingly smaller with every generation of trained GANs and that this smaller difference results primarily from VOT in the \#sTV sequences gradually becoming longer.  Table \ref{rawmes} gives raw means of VOT durations across the generations; Figure \ref{evolutionGANviolin} illustrates the distribution. While it appears that the average duration of VOT in the \#TV condition does not change substantially across the generations, VOT duration in the \#sTV condition substantially lengthens: it reaches   43.31ms  in \textsc{Gen4} (compared to 25.36ms in TIMIT). 

\begin{table}[ht]
\centering
\begin{tabular}{lccccc}
     \hline\hline
      &  \multicolumn{4}{c}{VOT in ms}   \\
 &  \multicolumn{2}{c}{\textbf{\#TV}}  &   \multicolumn{2}{c}{\textbf{\#sTV}} \\
 & Mean & SD& Mean& SD\\ 
  \hline
  \textsc{Gen0} & 59.33 & 20.97 & 25.36 & 8.76 \\ 
   \textsc{Gen1} & 61.83 & 24.76 & 38.85 & 18.44 \\ 
  \textsc{Gen2} & 58.42 & 22.44 & 40.40 & 19.74 \\ 
   \textsc{Gen3} & 58.88 & 24.52 & 42.56 & 20.63 \\ 
   \textsc{Gen4} & 62.41 & 26.36 & 43.31 & 18.30 \\

   \hline\hline
\end{tabular}

\caption{\label{rawmes}Means of  VOT durations in ms with standard deviation across the two conditions (\#TV and \#sTV) four generations (\textsc{Gen1--4}) and TIMIT (\textsc{Gen0}).}
\end{table}

\begin{figure}
\centering
\includegraphics[width=0.45\textwidth]{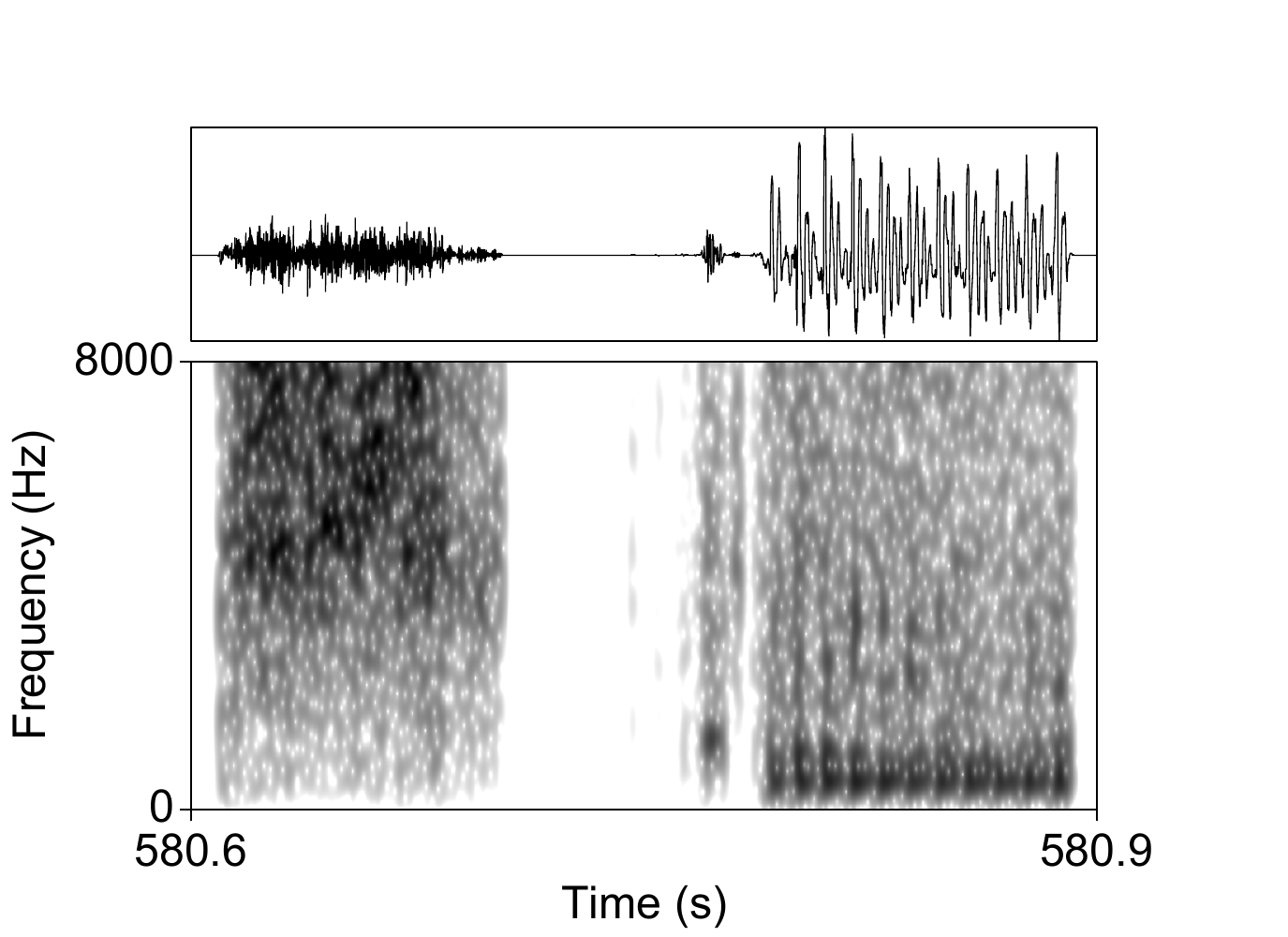}\includegraphics[width=0.45\textwidth]{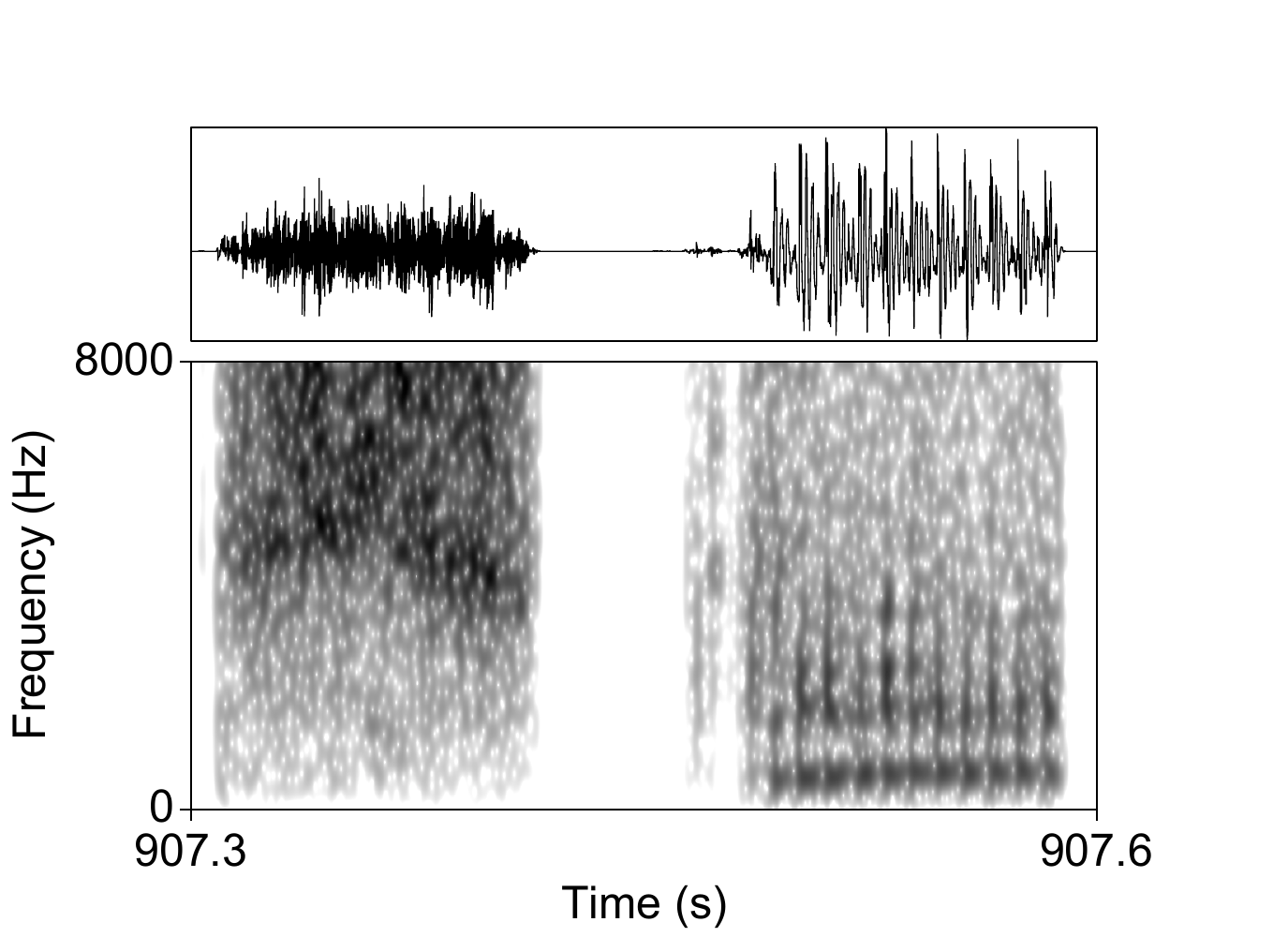}\\
\includegraphics[width=0.45\textwidth]{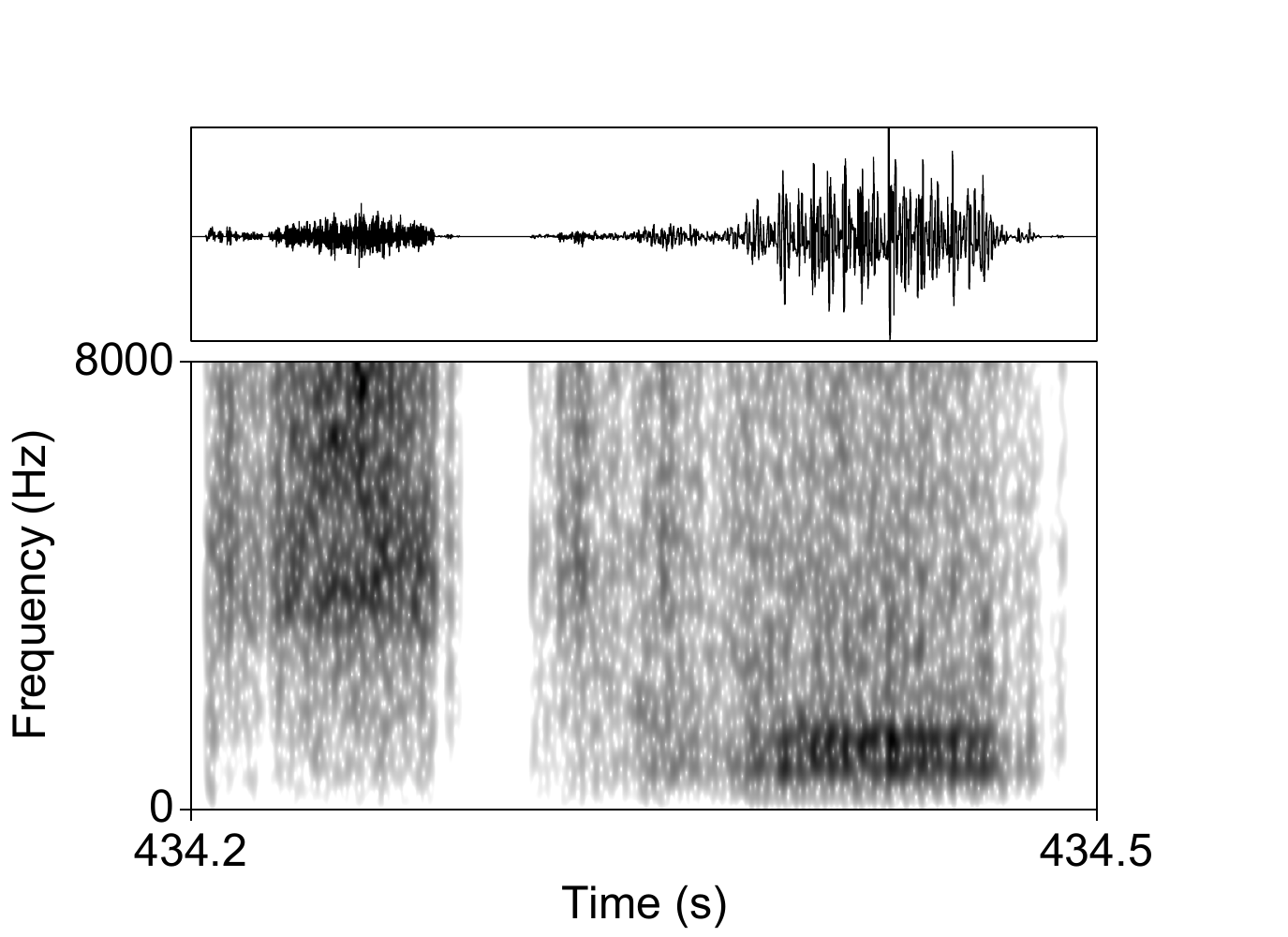}\includegraphics[width=0.45\textwidth]{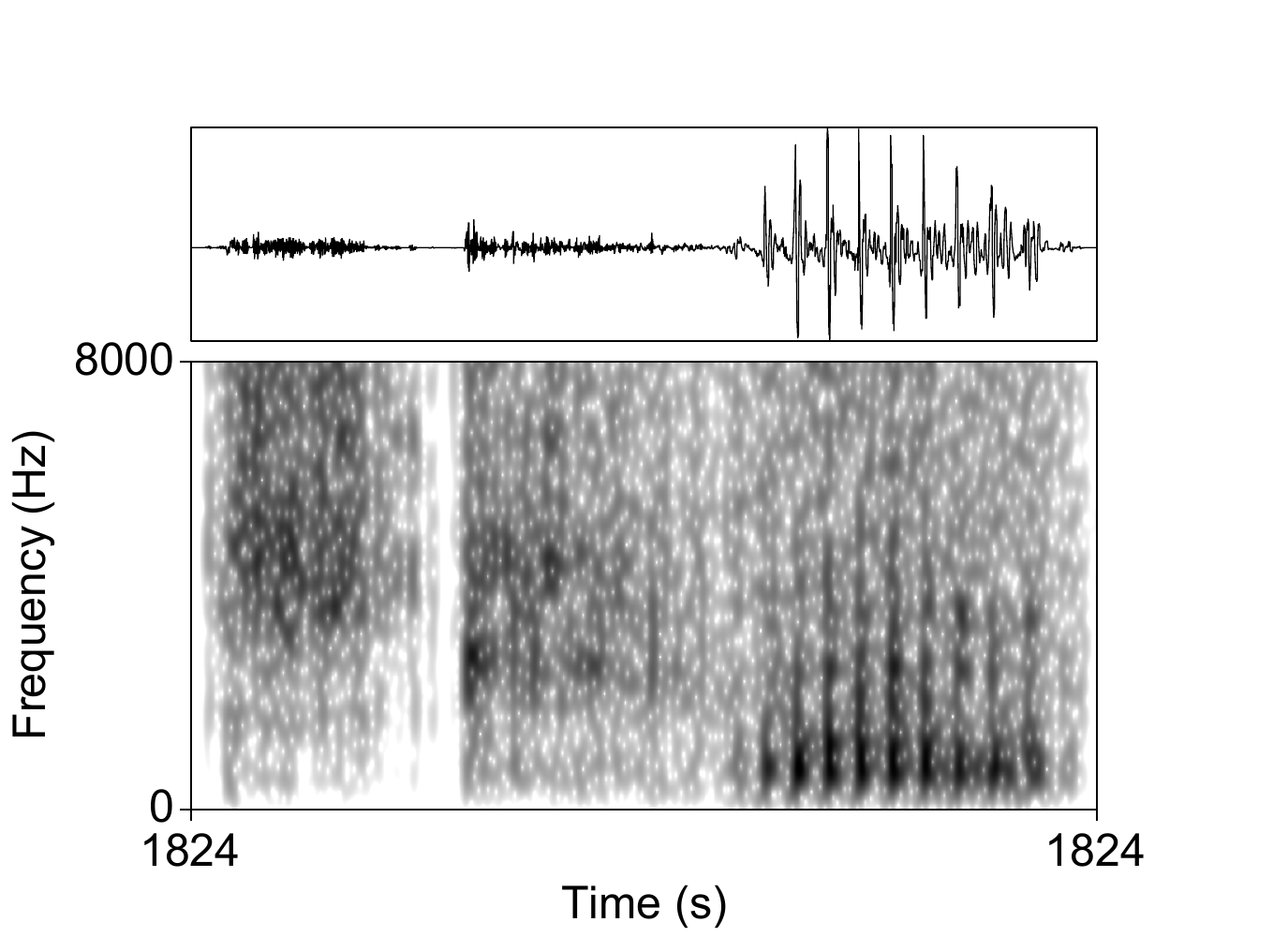}
\caption{\label{12255123} Waveforms and spectrograms (0-8,000 Hz) of four \#sTV sequences from \textsc{Gen1}, two with short VOT (upper row) and two with long VOT (lower row). The generator only outputs waveforms, spectrograms are provided only for the analysis.}
\end{figure}

The data distribution in the first generation of networks, \textsc{Gen1}, suggest that the network does learn the conditional distribution, but  imperfectly so, as the network outputs data in which the VOT duration in the \#sTV condition is longer than any VOT duration in this condition in the training data (see also \citealt{begus19}). Figure \ref{12255123} illustrates two outputs of \textsc{Gen1} with short VOTs (20.2 ms in  the left and 27.1 ms in the right sequence) and two outputs in which the VOTs are longer than any in the training data (82.4 ms in the left and 84.2 ms in the right). 

\begin{figure}
\centering
\includegraphics[width=0.45\textwidth]{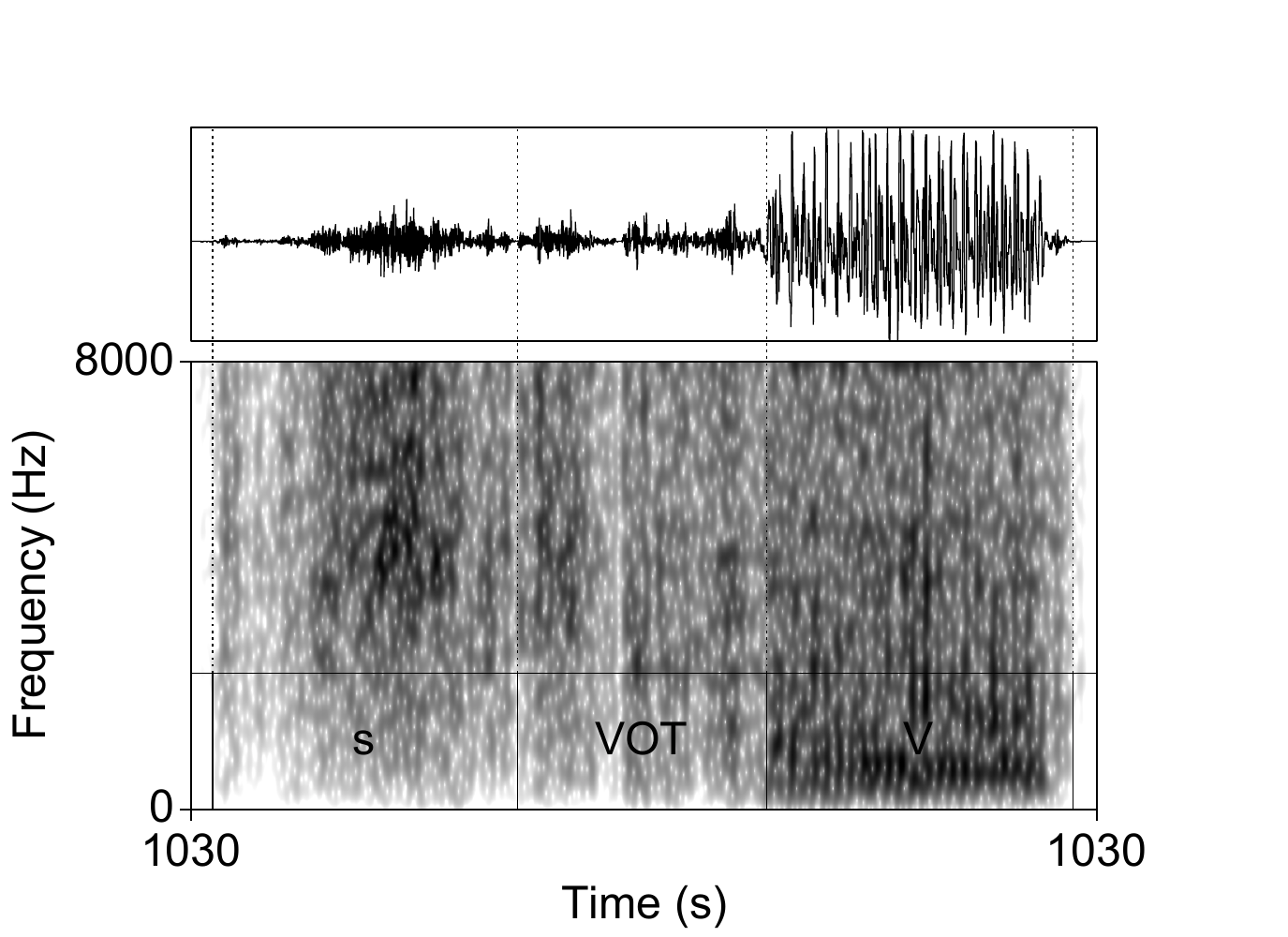}\includegraphics[width=0.45\textwidth]{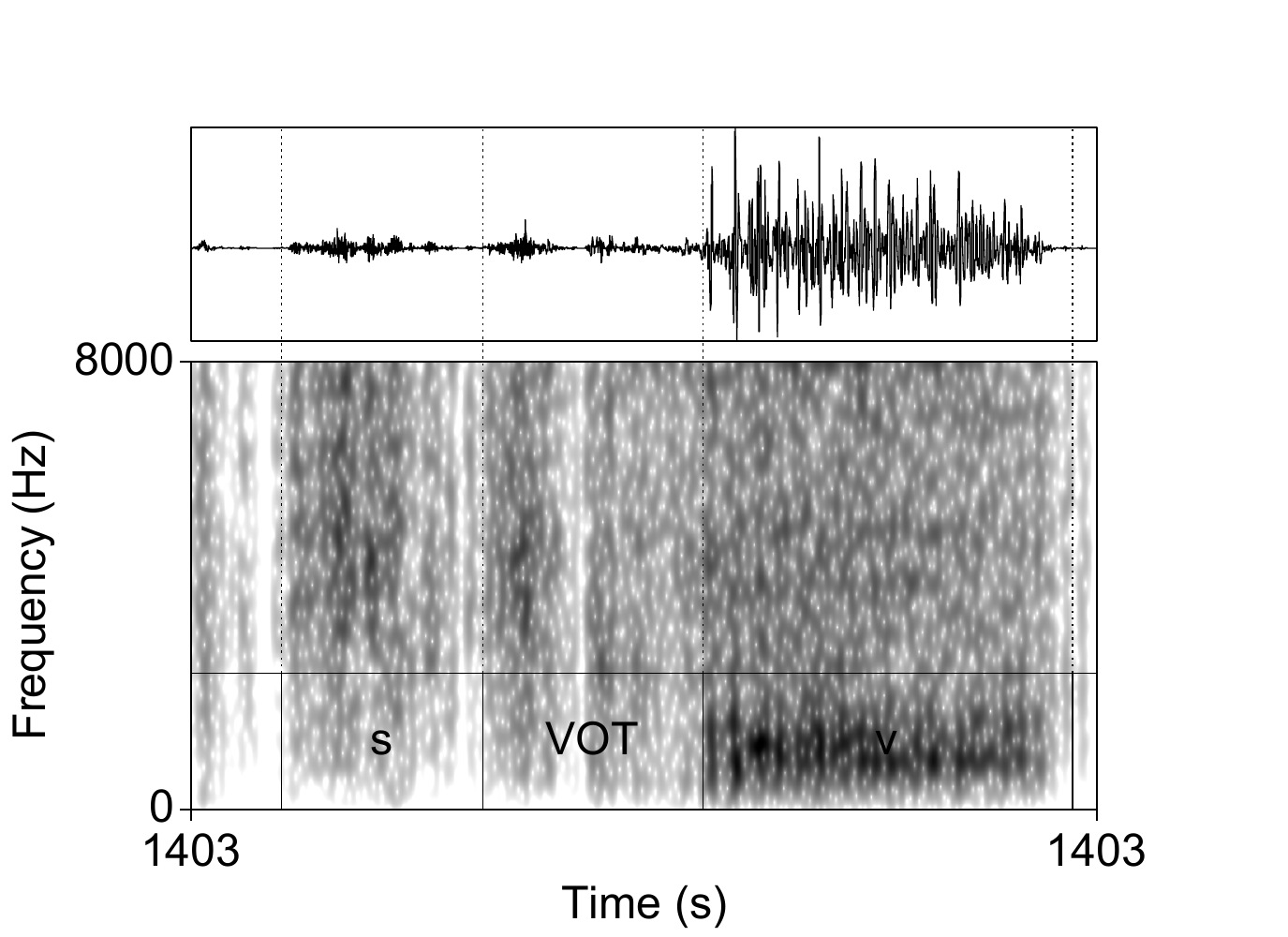}
\caption{\label{12246_1234} Waveforms and spectrograms (0--8,000 Hz) of two \#sTV sequences with typically long VOTs from \textsc{Gen4}. The network outputs only waveforms: spectrograms and annotations are given here for analysis.}
\end{figure}

By \textsc{Gen4}, outputs with short VOT become exceedingly rare, while outputs with long VOT, longer than in the TIMIT training data from \textsc{Gen0}, become very frequent. Figure \ref{12246_1234} illustrates this distribution with two examples of \#sTV sequences. The VOT duration at this point is long and the closure is almost entirely absent from the output. 

To test the significance of the difference in VOT durations across different generations, the data is fit to a Gamma regression model with log link (in \citealt{r}).\footnote{Gamma regression provides a better fit than an inverse Gaussian model (AIC $=-33010.92$ vs.~$-31925.08$).} The dependent variable is duration of VOT in ms. The model has two predictors: \textsc{Structure} (treatment-coded with \#sTV as the reference level) and \textsc{Generation} (treatment-coded with \textsc{Gen1} as the reference level) and an interaction between the two predictors. The AIC value of the model with the interaction is higher compared to the model without the interaction ($-33142$ vs.~$-32760$), which is why the interaction and the two predictors are kept in the final model. 

The model reveals several distributions in the data that are relevant to the questions posed by this paper. First, the model suggests that the difference in VOT duration between the \#TV and \#sTV conditions are significant for \textsc{Gen1}: $\beta -0.465,z=-11.03, p<0.0001$. In other words, the network learns to output shorter VOT in \textsc{Gen1} in the \#sTV condition compared to  the \#TV condition, which means that the network learns the conditional distribution of VOT duration (see also \citealt{begus19}).

While the network learns to output shorer VOT in the \#sTV condition, the learning is imperfect. This is suggested by the interaction term \textsc{Structure}:\textsc{Generation} which is a significant predictor (AIC = $-33011$ vs. $-32764$). Crucially, the difference in VOT duration between the two conditions (\#TV and \#sTV) is significantly greater in the TIMIT database compared to \textsc{Gen1} (trained on TIMIT): $\beta=-0.385, z=-8.44, p< 0.0001$. Similarly, this difference is significantly greater in \textsc{Gen1} compared to \textsc{Gen3}:  $\beta= 0.14,z=   2.36,p=   0.018$ (not significant for \textit{Gen4}; for other coefficients, see Table \ref{gammareg}). These results suggest that the Generator learns the allophonic distribution imperfectly with each generation. The difference in VOT duration between the two conditions is substantially and significantly smaller in the first generation (\textsc{Gen1}) compared to human training data (TIMIT), but this difference gets even smaller when further generations learn from each other (although the difference is smaller for later generations). In other words, the iterative learning results in increasingly smaller distinction in VOT duration between the \#TV and \#sTV conditions.

The absolute VOT durations in the \#sTV condition are significantly longer in the first generation trained on TIMIT (\textsc{Gen1}) compared to TIMIT ($\beta= -0.426, z=-12.39, p < 0.0000$) and significantly longer in the third and fourth generations compared to the first (\textsc{Gen1}): $\beta=0.091, z=2.17 p=0.030$ for \textsc{Gen3} and $\beta =0.109,z=2.40,p= 0.016$ for \textsc{Gen4}. The differences between \textsc{Gen1} and \textsc{Gen3} and \textsc{Gen4} are smaller in magnitude compared to the difference between \textsc{Gen0} and \textsc{Gen1}. 

\begin{figure}
\centering
\includegraphics[width=0.6\textwidth]{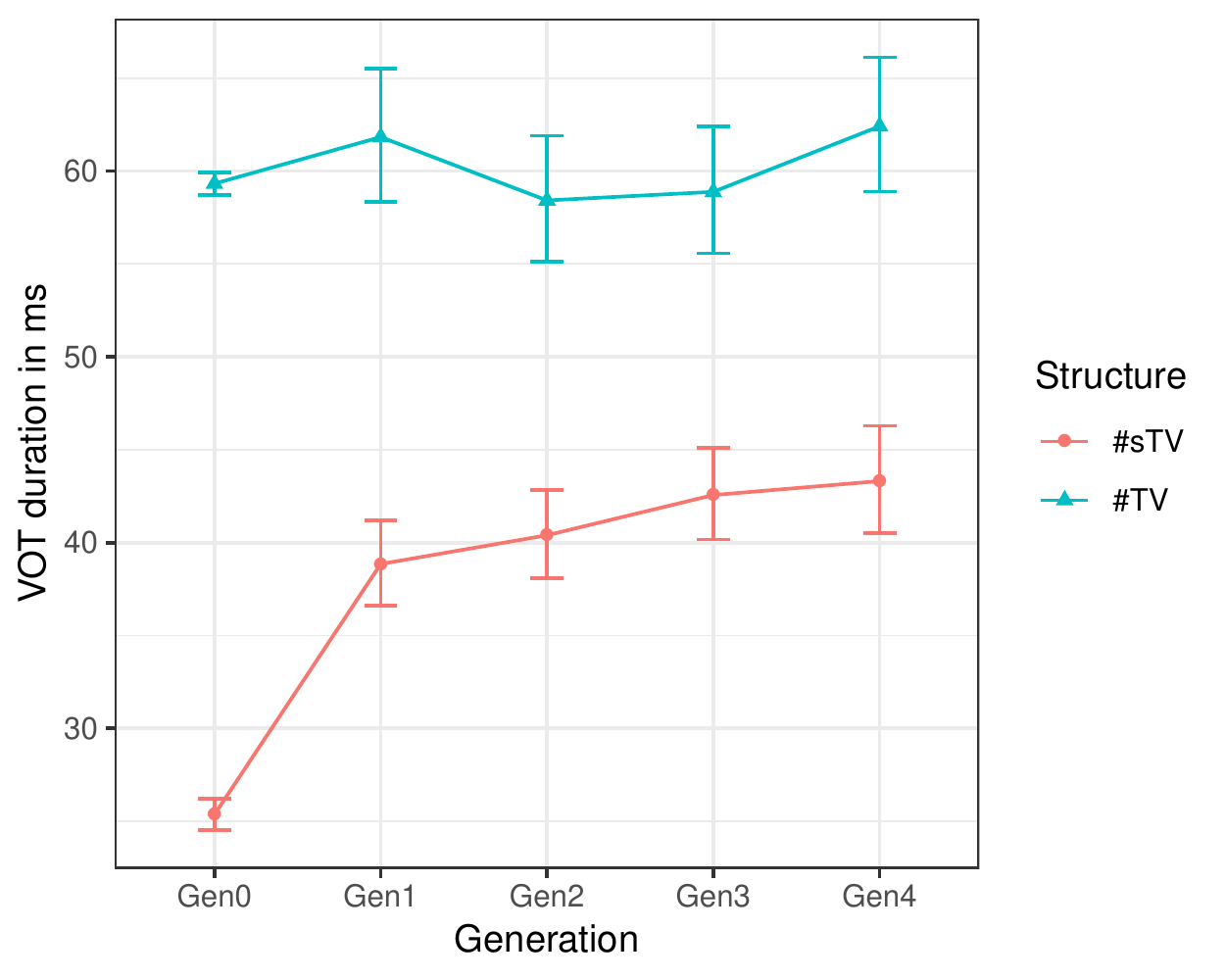}
\caption{\label{evolutionGAN} Predicted values based on the Gamma regression model in Table \ref{gammareg}.}
\end{figure}

In sum, there is a trend for the difference in VOT duration between the \#TV and \#sTV sequences to become smaller with each generation and especially for VOT in the \#sTV condition to become longer. Figure \ref{evolutionGAN} shows a clear pattern: while the VOT duration in the \#TV condition remains relatively constant across the generations, the VOT duration in the \#sTV condition rises steadily with each generation.  

Not only are the VOT durations of \#TV relatively stable across generations, they also do not differ from the original distribution in the training data (TIMIT). Post-hoc tests of the difference in the absolute VOT duration in the \#TV condition across the four generations suggest that VOT remains stable across generations. None of the pairwise comparisons between \textsc{Gen0} (TIMIT),  \textsc{Gen1}, \textsc{Gen2} and \textsc{Gen3} and \textsc{Gen4} are significant (see Table \ref{posthoc}). This suggests that the first generation of the Generator networks (as well as the subsequent generations) learn the  VOT durations in the \#TV condition from the training data relatively faithfully and that this duration remains stable in subsequent generations (unlike in the \#sTV condition).

\begin{table}[ht]
\centering
\begin{tabular}{rrrrr}
  \hline\hline
 & Estimate & Std. Error & t value & Pr($>$$|$t$|$) \\ 
  \hline
  
  (Intercept)=Gen1,\#sTV & -3.248 & 0.030 & -107.95 & 0.0000 \\ 
\#TV & 0.465 & 0.042 & 11.03 & 0.0000 \\ 
Gen0 & -0.426 & 0.034 & -12.39 & 0.0000 \\ 
Gen2 & 0.039 & 0.042 & 0.92 & 0.3567 \\ 
Gen3 & 0.091 & 0.042 & 2.17 & 0.0304 \\ 
Gen4 & 0.109 & 0.045 & 2.40 & 0.0164 \\ 
\#TV:Gen0 & 0.385 & 0.046 & 8.44 & 0.0000 \\ 
\#TV:Gen2 & -0.096 & 0.059 & -1.61 & 0.1073 \\ 
\#TV:Gen3 & -0.140 & 0.059 & -2.36 & 0.0182 \\ 
\#TV:Gen4 & -0.099 & 0.062 & -1.61 & 0.1065 \\

  \hline\hline
\end{tabular}
\caption{\label{gammareg} Coefficients of the Gamma regression model. }
\end{table}

\begin{table}[ht]
\centering
\begin{tabular}{lrrrrr}
  \hline\hline
Contrast & Ratio & SE & df & z.ratio & p.value \\ 
  \hline
\multicolumn{6}{l}{\textsc{Structure} = \#TV }\\
gen1 / \textsc{Gen0} & 1.0422 & 0.0312 & Inf & 1.381 & 0.6402 \\ 
  \textsc{Gen1} / \textsc{Gen2} & 1.0584 & 0.0442 & Inf & 1.360 & 0.6537 \\ 
  \textsc{Gen1} / \textsc{Gen3} & 1.0501 & 0.0438 & Inf & 1.171 & 0.7680 \\ 
  \textsc{Gen1} / \textsc{Gen4} & 0.9907 & 0.0413 & Inf & -0.223 & 0.9994 \\ 
  \textsc{Gen0} / \textsc{Gen2} & 1.0155 & 0.0304 & Inf & 0.512 & 0.9862 \\ 
  \textsc{Gen0} / \textsc{Gen3} & 1.0075 & 0.0302 & Inf & 0.250 & 0.9991 \\ 
  \textsc{Gen0} / \textsc{Gen4} & 0.9506 & 0.0285 & Inf & -1.692 & 0.4391 \\ 
  \textsc{Gen2} / \textsc{Gen3} & 0.9922 & 0.0414 & Inf & -0.189 & 0.9997 \\ 
  \textsc{Gen2} / \textsc{Gen4} & 0.9361 & 0.0391 & Inf & -1.583 & 0.5084 \\ 
  \textsc{Gen3} / \textsc{Gen4} & 0.9435 & 0.0394 & Inf & -1.394 & 0.6313 \\ 
  \hline\hline
\multicolumn{6}{l}{{\footnotesize P value adjustment: tukey method for comparing a family of 5 estimates}}\\

\multicolumn{6}{l}{{\footnotesize Tests are performed on the log scale}}\\

\end{tabular}
\caption{\label{posthoc} Post-hoc comparison of VOT duration in the \#TV condition across the generations with Tukey adjustment based on the Gamma regression model. }
\end{table}

\begin{figure}
\centering
\includegraphics[width=1\textwidth]{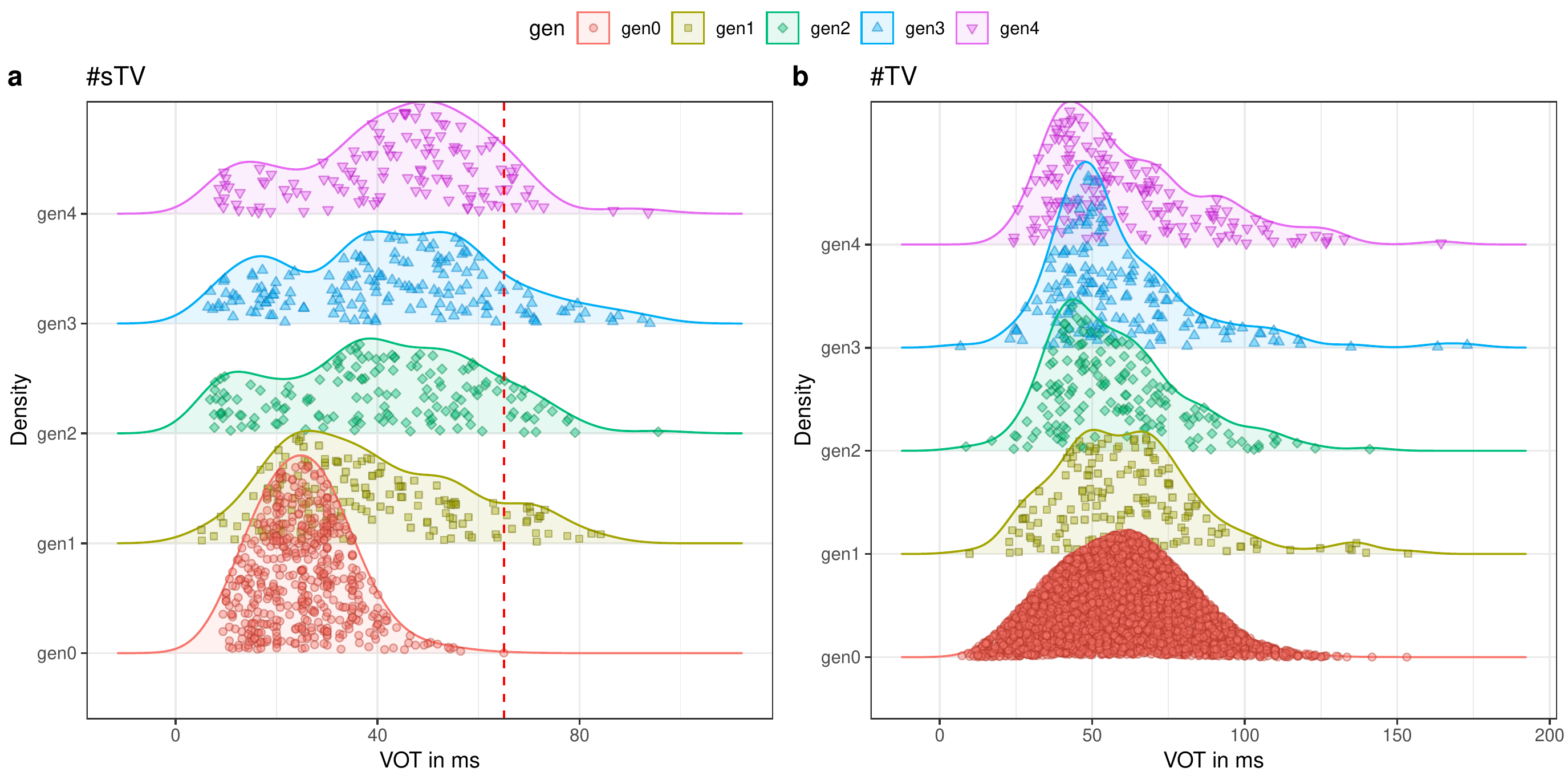}

\caption{\label{evolutionGANdensity} Density plots with data points of VOT duration in the \#sTV  (\textbf{a}) and \#TV (\textbf{b}) conditions across the four generations and TIMIT (\textsc{Gen0}). The red dashed line indicates the longest VOT duration in the \#sTV condition in the training data (TIMIT). }
\end{figure}

Even more informative for modeling sound change are density plots of VOT durations in the \#sTV condition in Figure \ref{evolutionGANdensity}. They reveal a clear trend, not only in overall duration, but that the peak VOT distribution is gradually increasing with each generation. VOT in the  TIMIT database (Gen0) is substantially shorter than in all other generations. The highest density of \textsc{Gen1} remains around the peak of TIMIT \textsc{Gen0}, while the distribution has substantially longer right tail with several data points in which the VOT is longer than any VOT duration in the training data. This suggests that the distribution in \textsc{Gen1} is learned, but imperfectly.  The next generations (Gen2, \textsc{Gen3}, and \textsc{Gen4}) show a gradual shift of the peak of the distribution towards longer VOT and additionally a trend towards a bimodal distribution. In all three generations, the outputs appear to have the majority distribution in what would be considered a long VOT that increasingly shifts towards even longer values. At the same time, the outputs also feature a lower peak around the original shorter VOT duration (closer to the TIMIT database). This sub-peak appears smaller in \textsc{Gen4} compared to \textsc{Gen3}. In the \#TV condition, the peaks are relatively stable across the four generations and TIMIT.

\subsection{Spectral properties}
\label{spectral}

As already mentioned, spectral properties of aspiration noise can differ across aspirated and unaspirated stops \citep{hussain20}. To test whether the change of VOT durations also entails changes of spectral properties of aspiration, we perform spectral analyses of the VOT duration. 
We are primarily interested in whether spectral properties change significantly across the four generations of iteratively trained networks when duration of the VOT is controlled for.

To test the stability of the aspiration's spectral properties across the four generations of neural networks, we perform spectral analysis of the aspiration noise measured at 10 intervals (from 10\% to 100\% of the duration). All 1,208 analyzed VOT durations (Table \ref{gentableprop}) across the four generations were thus analyzed for a total of 12,080 observations. For each annotated period of aspiration in the \#TV and \#sTV condition, we analyze Center of Gravity (COG), Standard Deviation, Skew, and Kurtosis of power spectra with 750-8,000 Hz filter (100Hz smoothing) using script by \citet{rentz17} in Praat \citep{boersma15}. Four generalized additive mixed effects models were fit for each of the four measured spectral properties (using the \emph{mgcv} package by \citealt{mgcv} following \citealt{soskuthy17}). The models include \textsc{structure} as a parametric term (reference \#sTV vs.~\#TV)  and thin-plate difference smooths for \textsc{Percent} of VOT duration (10--100\% in 10\% incremets), absolute \textsc{Duration} (full VOT duration in seconds), and \textsc{Generation} of neural networks (1-4, limited to 4 knots) and all two-way tensor product interactions. The model also includes random  smooths for each trajectory (1,208 total).

\begin{figure}
\centering
\includegraphics[width=0.24\textwidth]{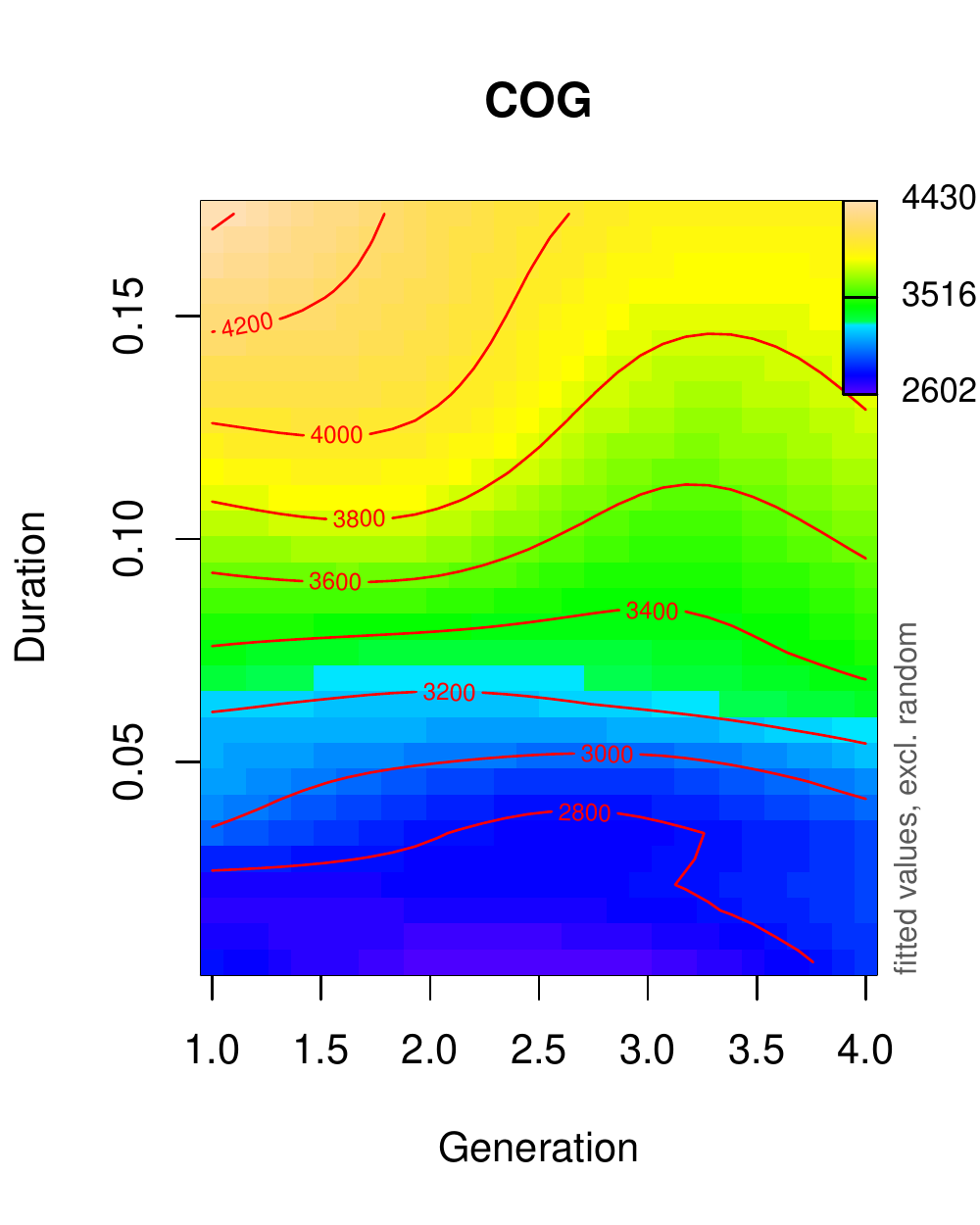}\includegraphics[width=0.24\textwidth]{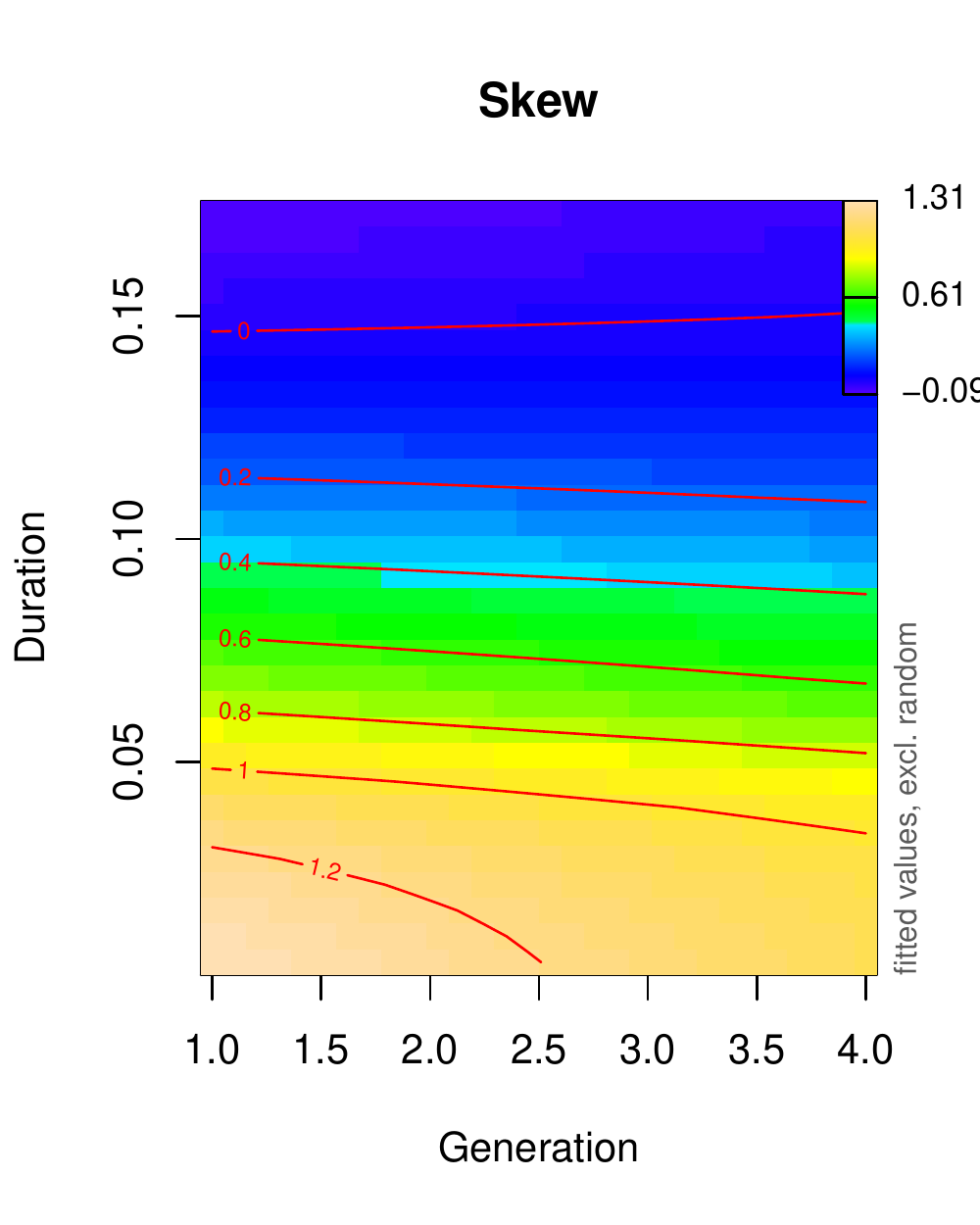}\includegraphics[width=0.24\textwidth]{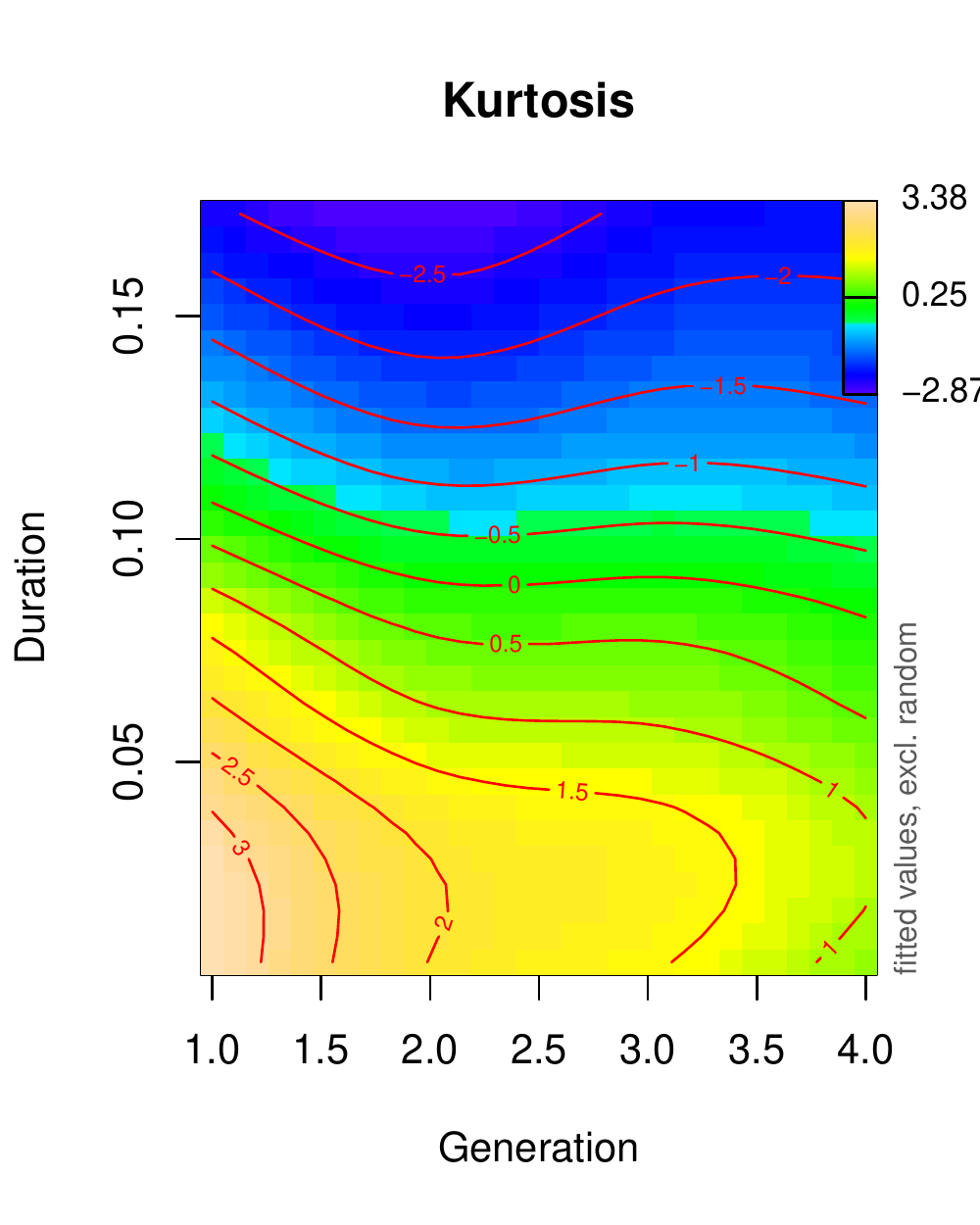}\includegraphics[width=0.24\textwidth]{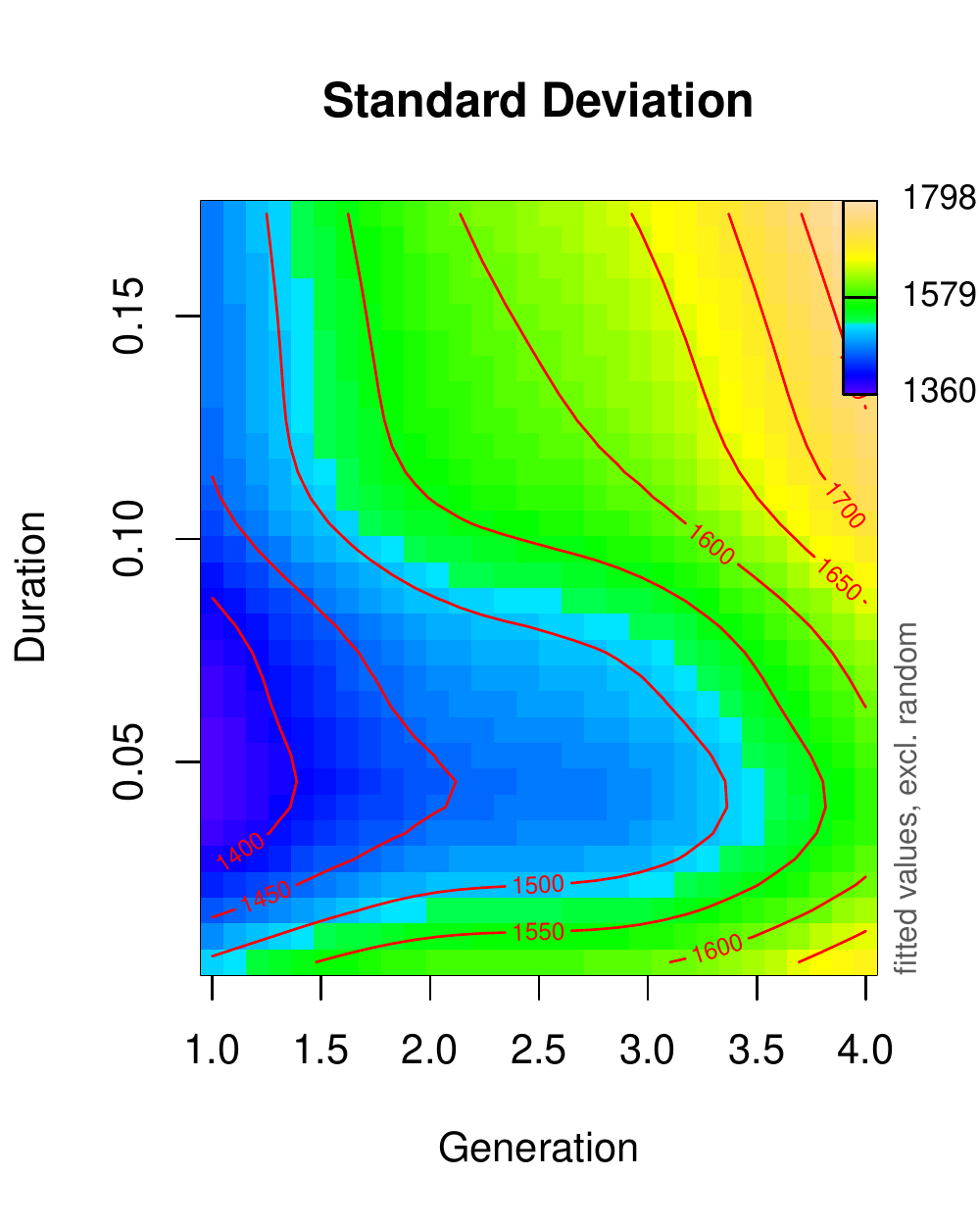}\\
\includegraphics[width=0.24\textwidth]{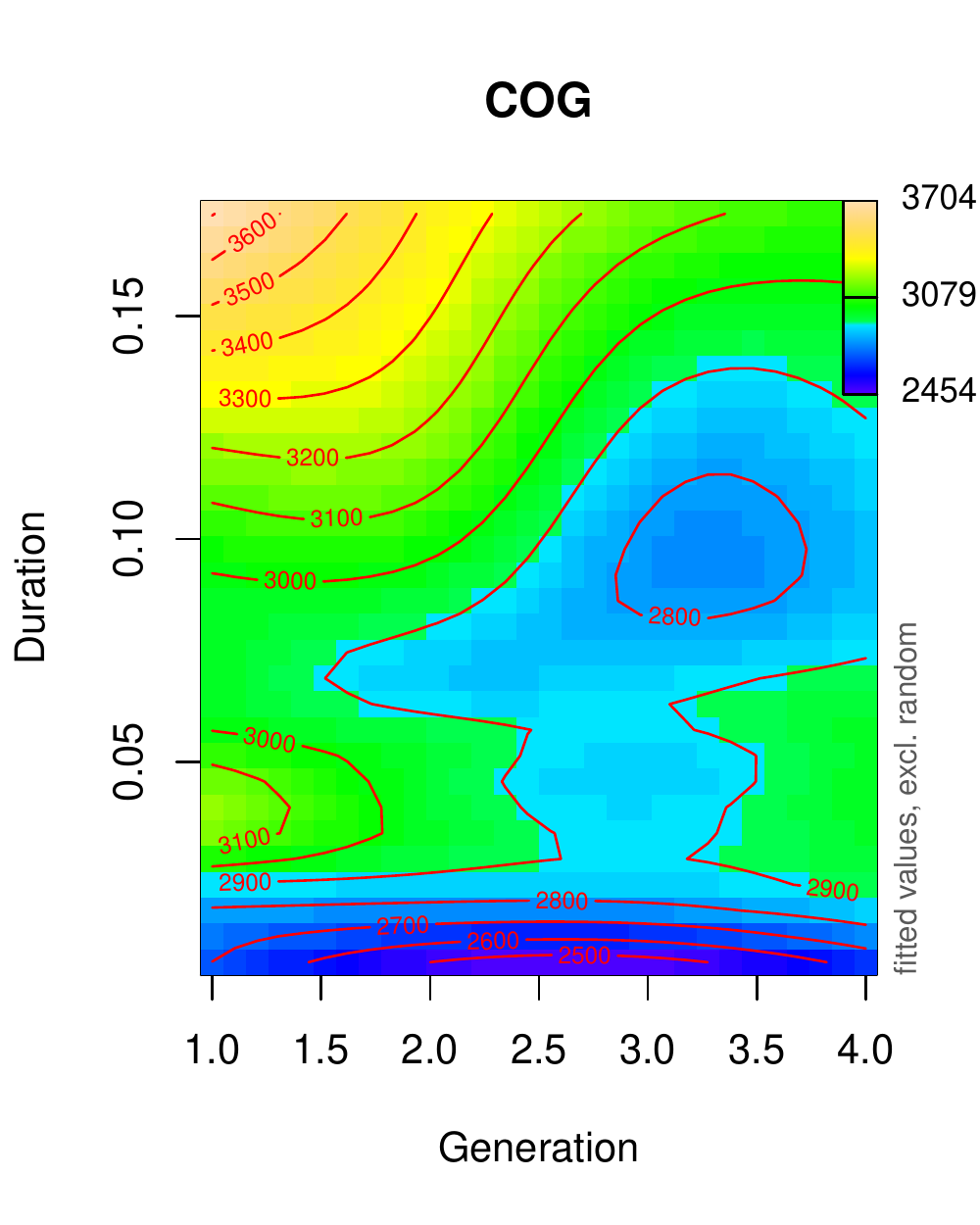}\includegraphics[width=0.24\textwidth]{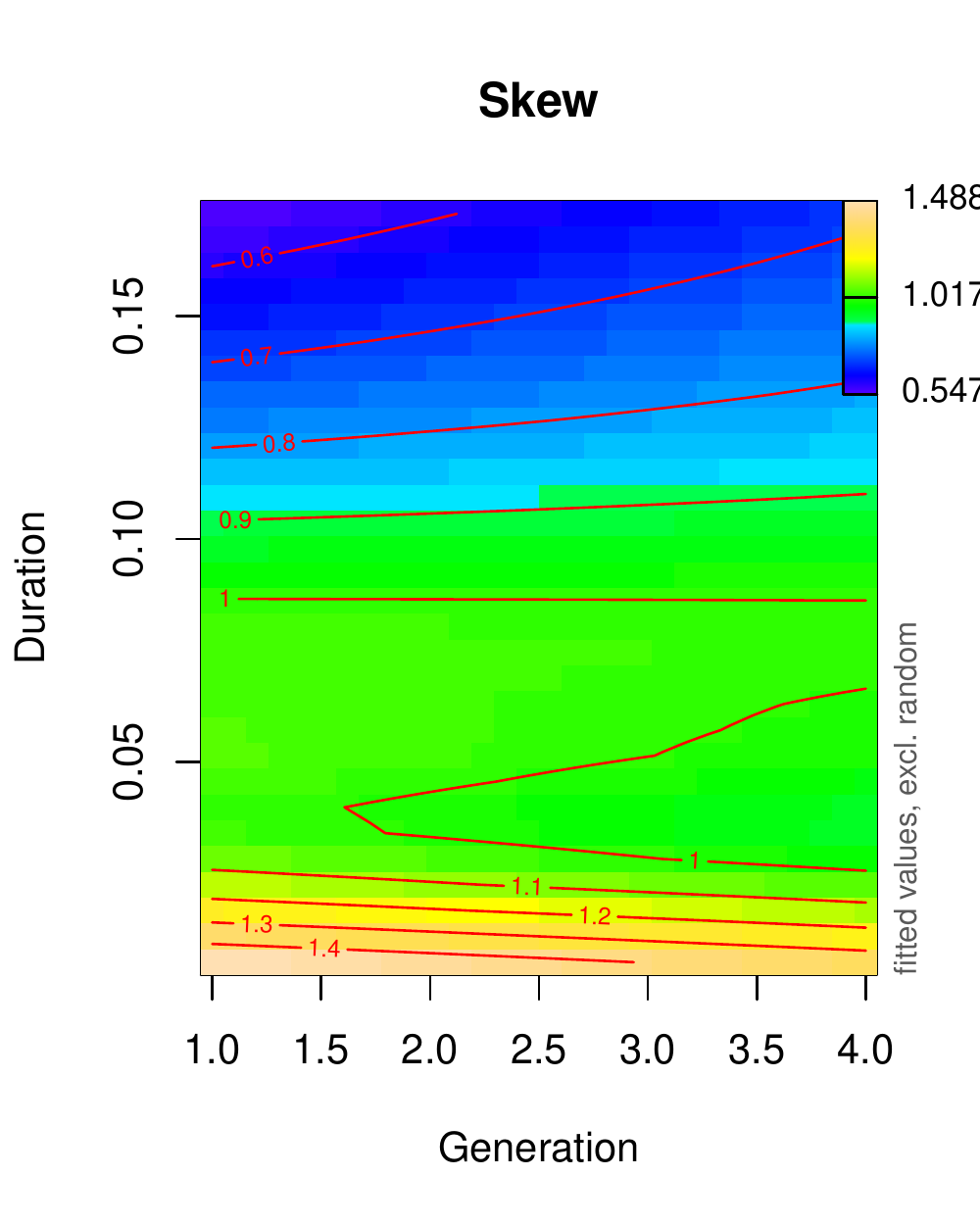}\includegraphics[width=0.24\textwidth]{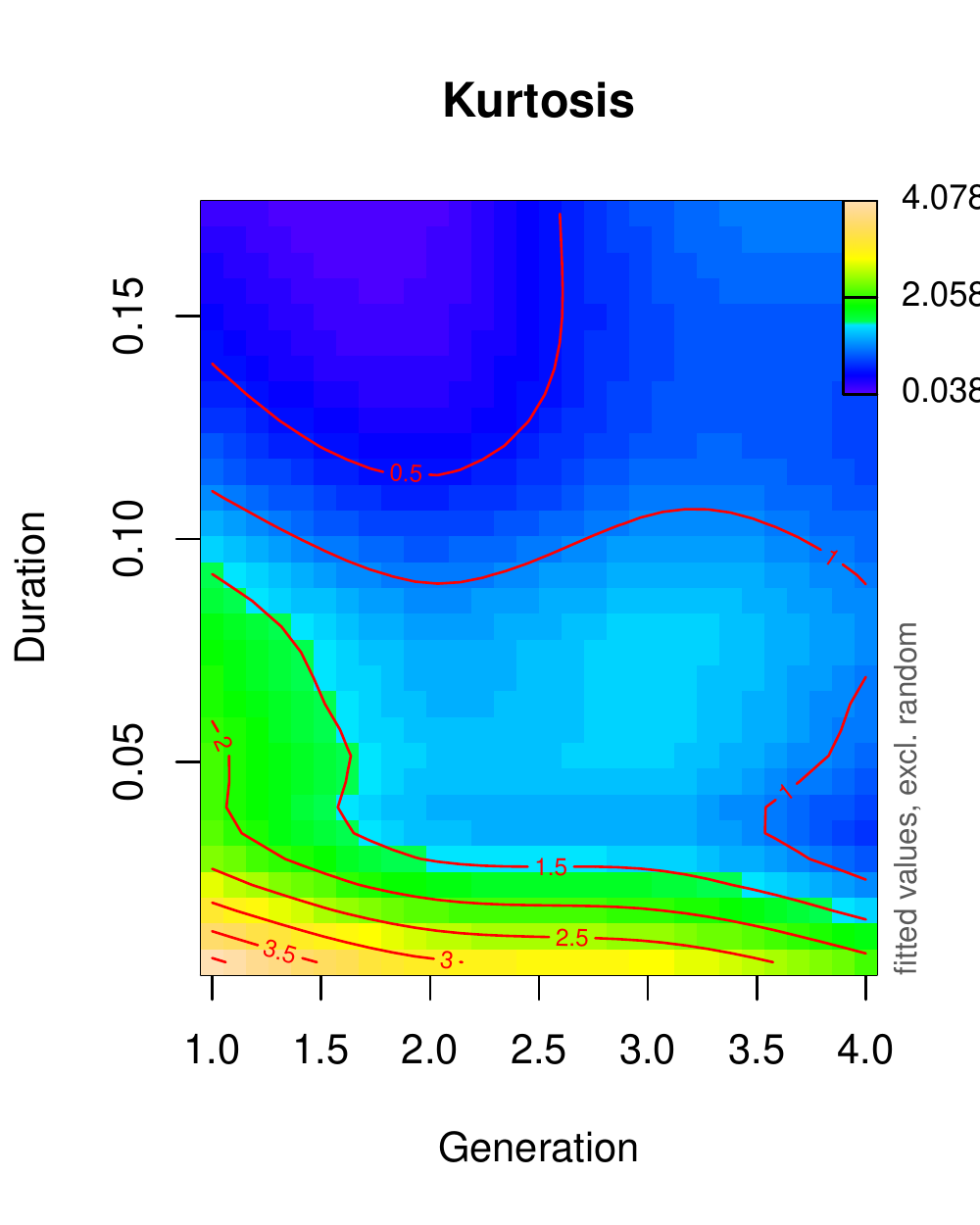}\includegraphics[width=0.24\textwidth]{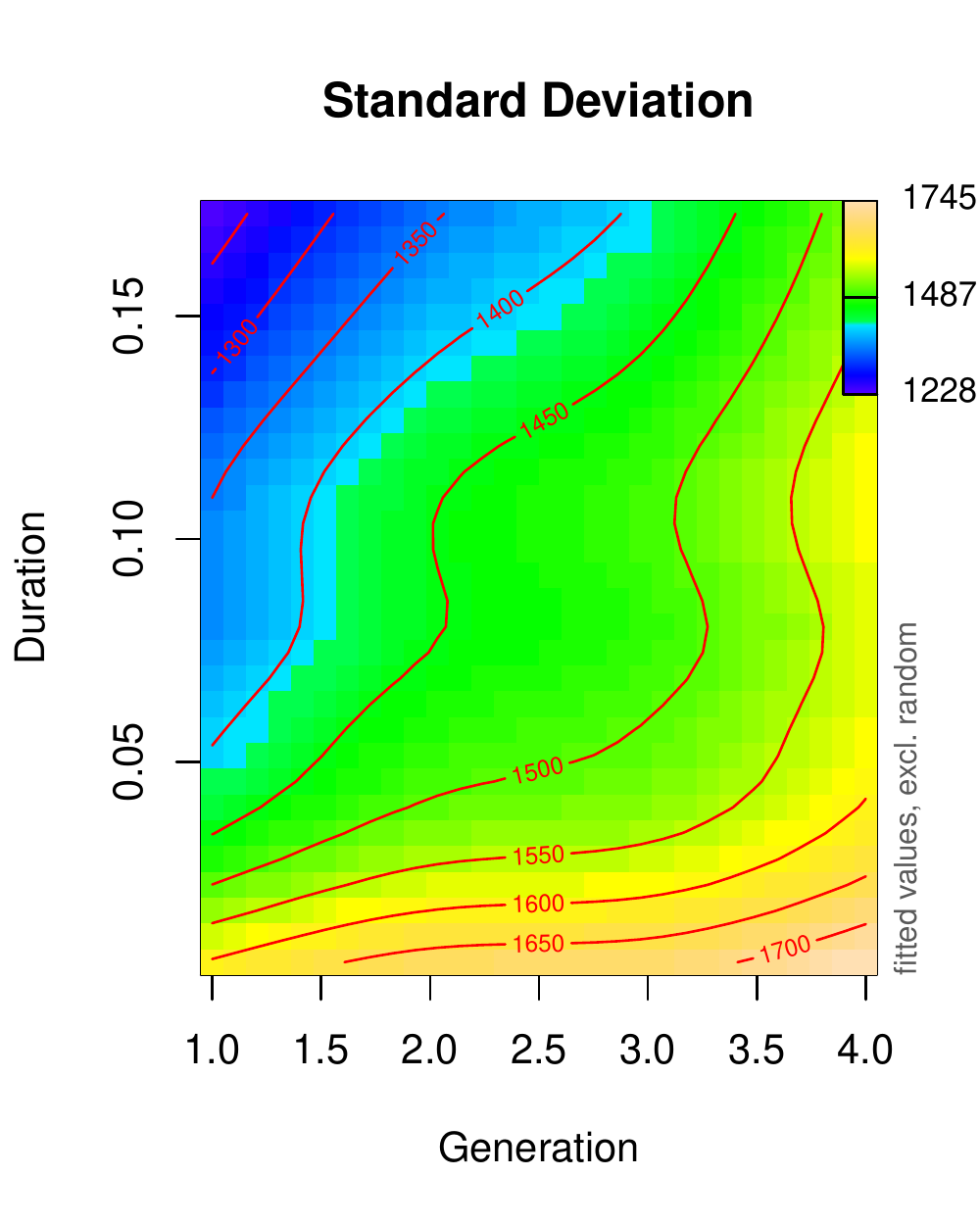}
\caption{\label{maps} Plots of nonlinear interactions between \textsc{Generation} and \textsc{Duration} of VOT for the four dependent variables, COG, Skew, Kurtosis, and Standard deviation at 10\% of VOT duration (upper row) and 50\% of VOT duration (lower row).}
\end{figure}

In addition to the change of VOT duration, the tensor product interaction between \textsc{Duration} of VOT and \textsc{Generation} is the most informative test for the question of whether acoustic/spectral properties significantly change across the four generations. Figure \ref{maps} plots the four spectral properties over the four generations and the various durations of VOT at 10\% and 50\% of VOT duration. All estimates of the four models are given in Tables \ref{COGtab.gam}, \ref{SKEWtab.gam}, \ref{KURTtab.gam}, \ref{SDEVtab.gam}. None of the estimates for the tensor product interaction \textsc{Duration}:\textsc{Generation} in the \#sTV condition are significant in the four models that measure COG, Skew, Kurtosis, and SD. The plots illustrate relatively constant values of the spectral values across the four generations. In other words, for the same levels of \textsc{Duration}, spectral properties do not change substantially across the iteratively trained generations of neural networks.

\section{Discussion}
\label{discussion}

Training deep convolutional networks in the GAN architecture in iterative learning tasks reveal several superficial similarities between language change and changes in acoustic outputs of the deep convolutional networks. First, the networks show relative stability in the proportion of \#TV and \#sTV sequences that they output. Human subjects  extend frequencies of phonological properties of the lexicon to novel test items in behavioral experiments \citep{hayes06,zymet18}.  This phenomenon is usually called \emph{frequency matching}. In Hungarian, for example, some roots can take a dative morpheme with the back vowel \textipa{[-nOk]}, while others roots take the same morpheme, but containing a front vowel \textipa{[-nEk]}. In a subclass of nouns, \textipa{[-nOk]} appears in 7.8\% cases in Hungarian lexicon. When speakers are asked to choose between \textipa{[-nOk]} and \textipa{[-nEk]} in novel unobserved words that resemble the actual class of nouns in which \textipa{[-nOk]} appears in 7.8\% cases, they closely match this frequency: they choose \textipa{[-nOk]} in 6.3\% of outputs in a behavioral experiment \citep{hayes06}.

Traces of such behavior are observed in our model: the proportion of [s]-initial sequences in the Generator's outputs is relatively stable across the four generations. This means that the network's output distribution closely matches training data. While the first network (\textsc{Gen1}) outputs slightly fewer \#sTV sequences compared to their representation in the TIMIT training data, two subsequent generations (\textsc{Gen2} and \textsc{Gen3}) do not diverge significantly in proportions of \#sTV sequences from their training data (see Figure \ref{evolutionGANproportionS}). The proportion is slightly, but significantly lower in \textsc{Gen4}, but 5.95\% of sequences stlll contain an [s]. The networks are thus relativesly successful in matching frequencies of the training data in the generated outputs, especially considering the fact that training is stopped after relatively few steps. While this result might seem trivial, it is conceivable that the considerably less frequent \#sTV sequences would be lost earlier with generations of networks --- parallel to the loss of the durational distribution of VOT (see below).

The networks also learn to output data that resembles speech and learn the conditional allophonic distribution in English --- deaspiration after [s] ([\textipa{"p\super hIt}] $\sim$ [\textipa{"spIt}]). In the first generation (\textsc{Gen1}) trained on TIMIT, the difference in VOT duration between the \#TV and \#sTV sequences is significant (see also \citealt{begus19}). The distribution peak of VOT durations in the \#sTV condition is in the first generation well-aligned with the original TIMIT distribution (Figure \ref{evolutionGANdensity}). Nevertheless, the distribution has a substantially longer right tail and the mean duration of VOT in \textsc{Gen1} is significantly longer than in TIMIT, as suggested by the Gamma regression model in Table \ref{gammareg}.

By \textsc{Gen4}, long VOT durations become so frequent that the distribution might be considered as not featuring the allophonic distribution at all. Plots of VOT distributions across generations (Figure \ref{evolutionGANdensity}) show  that the distribution peak gradually moves towards long VOT with each generation. While VOT is still shorter in the \#sTV compared to the \#TV condition in \textsc{Gen4}, the peak of VOT distribution is sufficiently long that most phonological analyses would likely analyze such a distribution as not involving an allophonic process. The slightly shorter VOT duration would likely be ascribed to the fact that the stop T is part of a consonant cluster (sT) and therefore automatically shorter. While VOT durations get increasingly longer with each generation, VOT durations in the more frequent \#TV condition remain stable and do not change significantly.

The generated outputs thus resemble both a phonetic gradual sound change and a phonological change ---  loss of a phonological rule. Peak distribution of VOT durations in the experiment gradually changes with each generation. This can be analyzed as superficially parallel to shifts in articulatory targets in a gradual phonetic sound change in natural language. The endpoint result, on the other hand, can be analyzed as a phonological rule loss. Rule loss in phonology \citep{kiparsky65,bermudez03} is a process whereby a language at stage A features a phonological rule and looses that rule at stage B.  When a distribution becomes automatic and does not require active control of the speakers, it is assumed that a process is part of an automatic phonetic distribution and not part of phonological computation. It is difficult to assess how actively the Generator controls VOT durations in \textsc{Gen4}, but it is likely the case that most linguistic analyses would treat such a minor difference in VOT duration as part of a phonetic rather than phonological shortening. So at least superficially, a phonological rule loss emerges in  the deep convolutional neural networks trained in iterative learning tasks: TIMIT (\textsc{Gen0}) features an allophonic rule of deaspiration of [s]-initial sequences, while at the stage of \textsc{Gen3} or \textsc{Gen4}, such rule is likely lost. 

The reason why the VOT duration in the [s]-condition gets gradually longer with imperfect learning whereas VOT durations in the \#TV sequences remain stable is parallel to phonological pressures in natural language. In our training data, \#TV sequences are significantly more frequent than \#sTV sequences: 4,930 vs.~533. This distribution actually reflects English lexicon: all sequences of the shape \#TV and \#sTV were sliced from TIMIT for the training data, but because \#TV is substantially more frequent in English lexicon, it is also more represented in TIMIT. The more frequent sequence in the training data is thus \#TV and therefore long VOT is the more general, less restricted variant (or in linguistic terms the ``unmarked'' elsewhere variant; see also Figure \ref{evolutionGANviolin}). This difference in training distributions affects the acquisition of  shorter  VOT durations in the \#sTV condition both in natural language acquisition as well as in deep convolutional networks trained in iterative learning tasks. However, while the property of the data --- the lower frequency of one condition --- is a linguistic property, the fact that imperfect learning in iterative tasks results in an apparent loss of a distribution is not linguistic (domain-specific) because the models that undergo learning have no pre-assumed linguistic information.

The results of the computational experiment also bear implications for models of language acquisition. The acquisition of VOT durations in human language learners undergoes several stages: first, VOT durations in voiceless stops are shorter than in adult targets in the \#TV condition, because long VOTs are more difficult to acquire \citep{gilbert77,macken80,lowenstein08,yang18}. Later, however, children produce long VOTs in \#TV sequences and in some cases even longer VOTs compared to the adult targets \citep{macken80}. The development of VOT durations in the \#sTV condition is even more intriguing. Crucially, children output substantially longer VOT durations compared to adult language in the \#sTV condition \citep{bond80} despite the fact that long VOTs are generally more difficult to acquire \citep{gilbert77,lowenstein08,yang18}. In other words, children extend the long VOTs in the \#TV condition  to the \#sTV condition and then need to acquire the shortening of VOT in the \#sTV condition. In fact, they often fail to shorten VOTs in the  \#sTV condition which results in long VOTs in \#sTV sequences \citep{bond80}. Just like in training data, \#sTV sequences are less frequent, which is likely the reason for the unexpected long VOTs in the \#sTV condition during acquisition. 

Both deep convolutional networks and language-acquiring children thus need to acquire the shortening of VOT in the more limited and less frequent [s]-condition. Failure to do so results in a long VOT duration in the \#sTV sequences. Imperfect learning that results in long VOT in the \#sTV sequences is attested in both language-acquiring children, as suggested by \citet{bond80}, and deep convolutional networks, as suggested by the computational simulations in this paper. Imperfect acquisition over time can accumulate in gradual shifts in phonetic targets \citep{baker08} which, at the endpoints, results in a phonological rule loss. Peaks of distributions of VOT duration in the four generations gradually move from short towards long VOT, and the end-result contains predominantly long VOT durations in the \#sTV conditions. Imperfect acquisition in the GAN networks is catalyzed by the fact that we train the networks for a relatively few training steps (less than 13,000).  The magnitude of imperfect acquisition is likely smaller in language-acquiring children. This is likely the reason for why appearance of rule loss emerges only after four generations of deep convolutional networks in our model, but substantially more speakers are usually needed for emergence of sound change. With less imperfect learning, the deep convolutional networks would likely take longer to achieve the same effect.  The principle behind the change nevertheless remains even if the acquisition is less imperfect than in our computational model.  

In addition to measuring VOT durations, we performed spectral analyses of aspiration noise that included COG, SD, skew, and kurtosis measured at every 10\% of aspiration duration. The results of non-linear regression models suggest that for equal VOT durations, spectral properties of frication noise do not change significantly across the generations.  In other words, when VOT duration is controlled for, the aspiration noise does not change significantly, which suggests that lengthening of the VOT is the primary change operating in the four generations of deep convolutional networks.

The author is unaware of any English varieties where such a rule loss would occur in this particular allophonic distribution. Other parallels, however, have been recorded in the literature on sound change in progress. For example, vowels in American English are nasalized before nasal stops.  In a word like \emph{pen} [\textipa{"p\super h\~En}], the vowel /\textipa{E}/ is often produced with nasalization ([\textipa{\~E}]), although this distribution is allophonic rather than contrastive (parallel to our allophonic rule of deaspiration). This nasalization results from coarticulation of nasalization: the gesture that causes nasality (raising of the velum) onsets early during the production of the vowel in anticipation of the nasal target of the following nasal stop. Like in our training data, the nasal variant of the vowel (e.g. [\textipa{\~E}]) is less frequent and more contextually limited compared to the oral vowel (e.g. [\textipa{E}]).  In some varieties, in fact, a sound change in progress has been reported, whereby nasalization coarticulation gets progressively shorter with generations of speakers \citep{zellou14,tamminga15} (\citealt{zellou14} also report a reversal of sound change). In other words, the targets for nasalization during the vowels onset gradually later with younger generations of speakers, which results in an oral [\textipa{E}] rather than nasal  [\textipa{\~E}]   in words like [\textipa{"p\super hEn}]. While this change is hypothesized to be driven by social factors \citep{tamminga15}, it is just as likely that phonological pressures in the form of the oral vowels being the more common variant play a role in the initiation of the change. 

\section{Conclusion}

The framework presented in this paper has several advantages for modeling language change. The networks are trained on raw acoustic inputs with no levels of abstraction or pre-extracted features. Deep convolutional networks in the GAN framework need to learn to produce data from random noise and they never fully replicate the data, but produce innovative linguistically interpretable data. This means their output data are not replicates, but innovative original outputs. As argued elsewhere \citep{begus19,begusCiw,begusLocal,begusIdentity}, the innovative outputs of the Generator are highly informative  and often replicate stages in language acquisition. In this paper we additionally argue that the innovative outputs can result in phonetic and phonological changes when trained in iterative learning tasks. 

The current model contains no articulatory information. While this is generally unideal, because human speech acquisition is highly influenced by articulators, it allows us to model language change as if only cognition-general mechanisms were involved in language acquisition and production. The results from the computational experiment suggest that both gradual change of targets that resembles phonetic change as well as phonological rule loss can emerge when deep convolutional networks are trained in iterative learning tasks without any articulatory information and without language-specific parameters in the model. In future work, we should be able to  compare results of the articulation-free model with models  containing articulatory information to get a better understanding of which properties of sound change can be derived from domain-general cognitive mechanisms and which properties of sound change require articulatory forces.

The current paper offers an initial step in a broader goal of modeling language's cultural evolution based on generations of deep convolutional networks trained on raw speech. Far more complex interactions between agents can be conceived in future work. For example, GANs  can be set to  communicate with and learn from each other in interactive ways already during the training. Further modeling of this kind should shed light onto one of the most widely studied but still poorly understood phenomenon in language --- sound change.

\subsubsection*{Acknowledgements}
This research was funded by a grant to new faculty at the University of Washington and the University of California, Berkeley. I would like to thank Sameer Arshad for slicing data from the TIMIT database.

\appendix

\section{Appendix}

\begin{table}
\centering
\begin{tabular}{rrrrr}
  \hline\hline
 & Estimate & Std. Error & z value & Pr($>$$|$z$|$) \\ 
  \hline
  (Intercept) = \textsc{Gen1}, \#TV & 7.5224 & 0.0259 & 290.4652 & $<$ 0.0001 \\ 
 	Gen0 & 0.9807 & 0.0317 & 30.9598 & $<$ 0.0001 \\ 
  \textsc{Gen2} & -0.0011 & 0.0366 & -0.0295 & 0.9764 \\ 
  \textsc{Gen3} & -0.0033 & 0.0366 & -0.0887 & 0.9293 \\ 
  \textsc{Gen4} & 0.0172 & 0.0365 & 0.4701 & 0.6383 \\ 
  \#sTV & -2.5051 & 0.0862 & -29.0761 & $<$ 0.0001 \\ 
  \textsc{Gen0}:\#sTV & 0.2805 & 0.0988 & 2.8395 & 0.0045 \\ 
  \textsc{Gen2}:\#sTV & 0.0142 & 0.1215 & 0.1172 & 0.9067 \\ 
  \textsc{Gen3}:\#sTV & 0.0422 & 0.1208 & 0.3494 & 0.7268 \\ 
  \textsc{Gen4}:\#sTV & -0.2553 & 0.1289 & -1.9806 & 0.0476 \\ 
 \hline\hline
\end{tabular}
\caption{\label{gentablepropLogReg}Coefficients of a negative binomial regression model of counts of \#sTV sequences in the generated data across the four generations and the TIMIT database (\textsc{Gen0}).}
\end{table}

\begin{table}
\centering
\scalebox{0.8}{\begin{tabular}{lrrrr}
   \hline\hline
A. parametric coefficients & Estimate & Std. Error & t-value & p-value \\    \hline
  (Intercept) = \#sTV & 2851.0891 & 21.9554 & 129.8580 & $<$ 0.0001 \\ 
 \#TV & 110.3979 & 29.9005 & 3.6922 & 0.0002 \\ 
   \hline
B. smooth terms & edf & Ref.df & F-value & p-value \\ \hline
  s(\textsc{Generation}) = \#sTV& 2.5277 & 2.5652 & 8.0885 & 0.0006 \\ 
  s(\textsc{Generation}):\#TV & 1.0006 & 1.0007 & 2.4167 & 0.1201 \\ 
  s(\textsc{Percent}) = \#sTV& 4.1403 & 5.0016 & 58.7212 & $<$ 0.0001 \\ 
  s(\textsc{Percent}):\#TV & 5.6734 & 6.7431 & 21.2203 & $<$ 0.0001 \\ 
  s(\textsc{Duration}) = \#sTV& 3.0980 & 3.1985 & 13.7967 & $<$ 0.0001 \\ 
  s(\textsc{Duration}):\#TV & 4.1206 & 4.2611 & 2.2167 & 0.0624 \\ 
  ti(\textsc{Percent},\textsc{Generation}) = \#sTV& 5.6734 & 7.3919 & 1.3457 & 0.2180 \\ 
  ti(\textsc{Percent},\textsc{Generation}):\#TV & 9.5814 & 12.3519 & 1.7253 & 0.0563 \\ 
  ti(\textsc{Duration},\textsc{Generation}) = \#sTV& 10.3210 & 10.8199 & 1.7353 & 0.0767 \\ 
  ti(\textsc{Duration},\textsc{Generation}):\#TV & 1.0038 & 1.0042 & 0.1402 & 0.7096 \\ 
  ti(\textsc{Duration},\textsc{Percent}) = \#sTV& 11.3020 & 12.4611 & 8.9140 & $<$ 0.0001 \\ 
  ti(\textsc{Duration},\textsc{Percent}):\#TV & 5.8098 & 6.8613 & 4.9689 & $<$ 0.0001 \\ 
  s(\textsc{Percent},\textsc{Trajectory},bs = ``fs'',m=1) & 3166.1177 & 10864.0000 & 1.2998 & $<$ 0.0001 \\ 
   \hline\hline
\end{tabular}}
\caption{Estimates of the generalized additive mixed effects model with COG as the dependent variable. The model includes the parametric term \textsc{structure} (treatment-coded with \#sTV as the reference level) and thin plate smooths for absolute \textsc{Duration}, \textsc{Percent} of VOT duration, and \textsc{Generation}, limited to 4 knots and random smooths for each trajectory. The model includes all two-way tensor product interactions.  } 
\label{COGtab.gam}
\end{table}

\begin{table}
\centering
\scalebox{0.8}{\begin{tabular}{lrrrr}
   \hline\hline
A. parametric coefficients & Estimate & Std. Error & t-value & p-value \\    \hline
  (Intercept) = \#sTV & 1.0995 & 0.0222 & 49.6085 & $<$ 0.0001 \\ 
 \#TV & -0.0753 & 0.0300 & -2.5134 & 0.0120 \\ 
   \hline
B. smooth terms & edf & Ref.df & F-value & p-value \\ \hline
  s(\textsc{Generation})= \#sTV & 1.0001 & 1.0001 & 0.7932 & 0.3732 \\ 
  s(\textsc{Generation}):\#TV & 1.0000 & 1.0000 & 0.2621 & 0.6087 \\ 
  s(\textsc{Percent}) = \#sTV& 3.5668 & 4.3585 & 45.4869 & $<$ 0.0001 \\ 
  s(\textsc{Percent}):\#TV & 4.0558 & 4.9625 & 19.2470 & $<$ 0.0001 \\ 
  s(\textsc{Duration}) = \#sTV& 3.3745 & 3.5492 & 11.0218 & $<$ 0.0001 \\ 
  s(\textsc{Duration}):\#TV & 3.0015 & 3.1647 & 0.7124 & 0.5352 \\ 
  ti(\textsc{Percent},\textsc{Generation}) = \#sTV& 1.0001 & 1.0001 & 3.8582 & 0.0495 \\ 
  ti(\textsc{Percent},\textsc{Generation}):\#TV & 4.8661 & 6.3659 & 2.4462 & 0.0264 \\ 
  ti(\textsc{Duration},\textsc{Generation}) = \#sTV& 1.0001 & 1.0001 & 0.4104 & 0.5218 \\ 
  ti(\textsc{Duration},\textsc{Generation}):\#TV & 1.0002 & 1.0002 & 0.0277 & 0.8679 \\ 
  ti(\textsc{Duration},\textsc{Percent}) = \#sTV& 11.8866 & 13.3683 & 10.9157 & $<$ 0.0001 \\ 
  ti(\textsc{Duration},\textsc{Percent}):\#TV & 4.0359 & 5.0538 & 1.4901 & 0.1627 \\ 
  s(\textsc{Percent},\textsc{Trajectory},bs = ``fs'',m=1) & 2364.7832 & 10864.0000 & 0.6582 & $<$ 0.0001 \\ 
   \hline\hline
\end{tabular}}
\caption{Estimates of the generalized additive mixed effects model with Skew as the dependent variable. The model includes the parametric term \textsc{structure} (treatment-coded with \#sTV as the reference level) and thin plate smooths for absolute \textsc{Duration}, \textsc{Percent} of VOT duration, and \textsc{Generation}, limited to 4 knots and random smooths for each trajectory. The model includes all two-way tensor product interactions. } 
\label{SKEWtab.gam}
\end{table}

\begin{table}
\centering
\scalebox{0.8}{\begin{tabular}{lrrrr}
   \hline\hline
A. parametric coefficients & Estimate & Std. Error & t-value & p-value \\    \hline
  (Intercept) = \#sTV& 1.7897 & 0.0804 & 22.2580 & $<$ 0.0001 \\ 
 \#TV  & 0.1426 & 0.1057 & 1.3487 & 0.1775 \\ 
   \hline
B. smooth terms & edf & Ref.df & F-value & p-value \\ \hline
  s(\textsc{Generation})= \#sTV & 2.7896 & 2.8498 & 9.8579 & $<$ 0.0001 \\ 
  s(\textsc{Generation}):\#TV & 1.7272 & 1.8634 & 0.7376 & 0.4462 \\ 
  s(\textsc{Percent})= \#sTV & 4.5849 & 5.6104 & 10.9902 & $<$ 0.0001 \\ 
  s(\textsc{Percent}):\#TV & 1.6770 & 2.0465 & 19.0525 & $<$ 0.0001 \\ 
  s(\textsc{Duration})= \#sTV & 4.1470 & 4.5475 & 7.7653 & $<$ 0.0001 \\ 
  s(\textsc{Duration}):\#TV & 1.0000 & 1.0000 & 0.7506 & 0.3863 \\ 
  ti(\textsc{Percent},\textsc{Generation})= \#sTV & 1.9587 & 2.2062 & 6.0669 & 0.0017 \\ 
  ti(\textsc{Percent},\textsc{Generation}):\#TV & 1.0002 & 1.0004 & 5.5611 & 0.0184 \\ 
  ti(\textsc{Duration},\textsc{Generation})= \#sTV & 1.0004 & 1.0005 & 2.5465 & 0.1106 \\ 
  ti(\textsc{Duration},\textsc{Generation}):\#TV & 1.0002 & 1.0003 & 0.3882 & 0.5333 \\ 
  ti(\textsc{Duration},\textsc{Percent})= \#sTV & 11.7809 & 13.5848 & 11.1060 & $<$ 0.0001 \\ 
  ti(\textsc{Duration},\textsc{Percent}):\#TV & 1.7229 & 1.9216 & 0.4682 & 0.5861 \\ 
  s(\textsc{Percent},\textsc{Trajectory},bs = ``fs'',m=1) & 1543.5132 & 10864.0000 & 0.2827 & $<$ 0.0001 \\ 
   \hline\hline
\end{tabular}}
\caption{Estimates of the generalized additive mixed effects model with Kurtosis as the dependent variable. The model includes the parametric term \textsc{structure} (treatment-coded with \#sTV as the reference level) and thin plate smooths for absolute \textsc{Duration}, \textsc{Percent} of VOT duration, and \textsc{Generation}, limited to 4 knots and random smooths for each trajectory. The model includes all two-way tensor product interactions. } 
\label{KURTtab.gam}
\end{table}

\begin{table}
\centering
\scalebox{0.8}{\begin{tabular}{lrrrr}
   \hline\hline
A. parametric coefficients & Estimate & Std. Error & t-value & p-value \\    \hline
  (Intercept) = \#sTV& 1506.5772 & 5.3036 & 284.0665 & $<$ 0.0001 \\ 
 \#TV  & -58.5769 & 6.9805 & -8.3915 & $<$ 0.0001 \\ 
   \hline
B. smooth terms & edf & Ref.df & F-value & p-value \\ \hline
  s(\textsc{Generation})= \#sTV & 2.9165 & 2.9493 & 54.3544 & $<$ 0.0001 \\ 
  s(\textsc{Generation}):\#TV & 1.9670 & 2.1560 & 10.5145 & $<$ 0.0001 \\ 
  s(\textsc{Percent}) = \#sTV& 6.5730 & 7.6910 & 9.5337 & $<$ 0.0001 \\ 
  s(\textsc{Percent}):\#TV & 3.4464 & 4.2477 & 5.0991 & 0.0003 \\ 
  s(\textsc{Duration})= \#sTV & 5.1565 & 5.7368 & 24.4827 & $<$ 0.0001 \\ 
  s(\textsc{Duration}):\#TV & 1.0003 & 1.0004 & 5.2345 & 0.0222 \\ 
  ti(\textsc{Percent},\textsc{Generation})= \#sTV & 14.1105 & 17.9717 & 2.6478 & 0.0002 \\ 
  ti(\textsc{Percent},\textsc{Generation}):\#TV & 1.3648 & 1.5431 & 1.4046 & 0.1664 \\ 
  ti(\textsc{Duration},\textsc{Generation})= \#sTV & 1.0005 & 1.0007 & 1.9439 & 0.1634 \\ 
  ti(\textsc{Duration},\textsc{Generation}):\#TV & 2.0627 & 2.3099 & 1.1559 & 0.4356 \\ 
  ti(\textsc{Duration},\textsc{Percent})= \#sTV & 6.4038 & 8.2036 & 6.0598 & $<$ 0.0001 \\ 
  ti(\textsc{Duration},\textsc{Percent}):\#TV & 1.9721 & 2.2224 & 1.1459 & 0.3054 \\ 
  s(\textsc{Percent},\textsc{Trajectory},bs = ``fs'',m=1) & 1226.7901 & 10864.0000 & 0.1875 & $<$ 0.0001 \\ 
   \hline\hline
\end{tabular}}
\caption{Estimates of the generalized additive mixed effects model with SD as the dependent variable. The model includes the parametric term \textsc{structure} (treatment-coded with \#sTV as the reference level) and thin plate smooths for absolute \textsc{Duration}, \textsc{Percent} of VOT duration, and \textsc{Generation}, limited to 4 knots and random smooths for each trajectory. The model includes all two-way tensor product interactions. } 
\label{SDEVtab.gam}
\end{table}

\clearpage

  \bibliographystyle{elsarticle-harv} 
   \bibliography{begusGANbib.bib}

\end{document}